\documentclass{article}

\usepackage[main,preprint]{neurips_2026}
\makeatletter
\renewcommand{\@noticestring}{Preprint. Accepted at NeurIPS 2026.}
\makeatother

\usepackage[utf8]{inputenc} %
\usepackage[T1]{fontenc}    %
\usepackage{hyperref}       %
\usepackage{url}            %
\usepackage{booktabs}       %
\usepackage{amsfonts}       %
\usepackage{nicefrac}       %
\usepackage{microtype}      %
\usepackage{xcolor}         %

\usepackage{graphicx}
\usepackage{subcaption}
\usepackage{amsmath}
\usepackage{amssymb}
\usepackage{mathtools}
\usepackage{amsthm}
\usepackage{colortbl}
\usepackage{enumitem}
\usepackage{multirow}
\usepackage{tabularx}
\usepackage{makecell}
\hypersetup{hidelinks,pdftitle={The Illusion of Diversity: Aligning LLM Exploration via Effective Entropy},pdfauthor={Xiaoliang Fu, Jiaye Lin, Yangyi Fang, Cong Qin, Binbin Zheng, Chaowen Hu, Zekai Shao}}
\AtBeginDocument{\raggedbottom}

\theoremstyle{plain}
\newtheorem{theorem}{Theorem}[section]

\theoremstyle{definition}

\theoremstyle{remark}

\newcommand{\E}{\mathbb{E}}

\title{The Illusion of Diversity: Aligning LLM Exploration via Effective Entropy}

\author{%
  \textbf{Xiaoliang Fu}\textsuperscript{1,*} \quad
  \textbf{Jiaye Lin}\textsuperscript{2,*,$\dagger$} \quad
  \textbf{Yangyi Fang}\textsuperscript{3,*} \quad
  \textbf{Cong Qin}\textsuperscript{4,*} \\
  \textbf{Binbin Zheng}\textsuperscript{5} \quad
  \textbf{Chaowen Hu}\textsuperscript{2} \quad
  \textbf{Zekai Shao}\textsuperscript{1} \\
  \textsuperscript{1}Fudan University \quad \textsuperscript{2}Meituan \quad
  \textsuperscript{3}Tsinghua University \\
  \textsuperscript{4}Peking University \quad
  \textsuperscript{5}University of Science and Technology of China \\
  {\textsuperscript{*}Equal contribution. \quad
  \textsuperscript{$\dagger$}Corresponding author: \texttt{linjiaye@meituan.com}}
}

\definecolor{improvecolor}{RGB}{45, 60, 145}

\definecolor{headerblue}{RGB}{44, 62, 80}
\definecolor{subheaderblue}{RGB}{235, 240, 245}
\definecolor{highlightcolor}{RGB}{230, 236, 245}

\begin{document}

\maketitle

\begin{abstract}
Reinforcement learning with verifiable rewards has shifted the alignment paradigm of large language models from subjective preference to objective correctness, yet a fundamental disconnect persists between exploration diversity and reasoning validity. Existing strategies predominantly rely on ``blind'' entropy metrics, e.g., token-level or global sequence-level entropy, which structurally fail to distinguish between effective reasoning and potential hallucinations. This misalignment creates an illusion of diversity, where high entropy tends to signal chaotic degeneration rather than valuable exploration. To bridge this gap, we introduce \textit{Effective Entropy}, a validity-aware entropy concept that quantifies diversity exclusively within the valid subspace and systematically filters out the noise of incorrect trajectories. Built on this concept, we instantiate VEPO, a simple \textbf{V}alidity-aware \textbf{E}ntropy \textbf{P}olicy \textbf{O}ptimization method that directly incorporates effective entropy into RLVR training. VEPO induces an adaptive attraction dynamic that drives up the probability of under-explored valid trajectories, a mechanism that matches the practical regime of large-scale autoregressive models where sequence probabilities remain extremely small. Experiments on six mathematical reasoning benchmarks with 1.5B and 7B DeepSeek-R1-Distill-Qwen models show consistent aggregate accuracy gains over GRPO and competitive repeated-sampling success. A direct evaluation with matched correct counts finds more distinct strategies on commonly solved MATH500 problems, supported by paired statistical tests and three-seed training comparisons.
\end{abstract}

\section{Introduction}

The alignment of Large Language Models (LLMs) has recently witnessed a paradigm shift from preference-based optimization to Reinforcement Learning with Verifiable Rewards (RLVR)~\citep{guo2025deepseek, lightman2023let, uesato2022solving, shao2024deepseekmath}. Unlike subjective human preferences, RLVR anchors model training in objective correctness by utilizing ground-truth signals, such as compiler outputs or mathematical answers, to verify reasoning chains. This shift has driven substantial progress in complex reasoning tasks, where algorithms such as Group Relative Policy Optimization (GRPO)~\citep{shao2024deepseekmath} show that simple statistical baselines can guide model improvement without the computational overhead of value networks.

\begin{figure*}[t]
\centering
\includegraphics[width=1.0\linewidth]{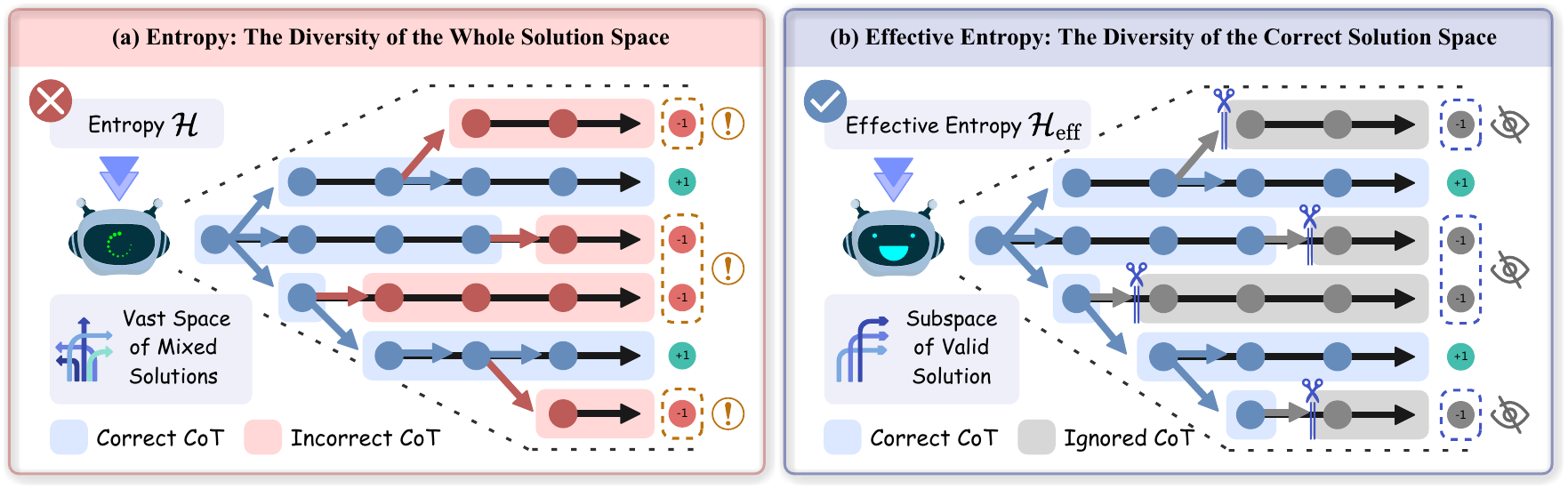}
\caption{Exploration in RLVR. \textbf{(a) Standard entropy:} Conventional entropy operates on the whole solution space. Since the invalid subspace is combinatorially dominant, blind maximization tends to disperse probability toward diverse errors. \textbf{(b) Effective entropy:} We measure diversity only within the valid subspace. Filtering out invalid trajectories provides a validity-aware view of exploration and motivates a new optimization direction for reasoning alignment.}
\vspace{-0.5em}
\label{fig:intro}
\end{figure*}

However, a disconnect persists between the goal of reasoning and the mechanism used to encourage exploration. Existing RLVR methods often rely on standard entropy to mitigate premature convergence and encourage diversity~\citep{williams1991function, eysenbach2021maximum, ahmed2019understanding, he2025skywork}. We argue that this reliance creates what we call \textit{the illusion of diversity}. As illustrated in Figure~\ref{fig:intro}, standard entropy is agnostic to validity and treats valid and invalid trajectories symmetrically. Because incorrect reasoning paths are combinatorially much more abundant than correct ones, blindly increasing entropy can disperse probability mass into the invalid region. As a result, high entropy may reflect chaotic degeneration rather than useful exploration.

To address this problem, we propose \textit{Effective Entropy} ($\mathcal{H}_{\mathrm{eff}}$), a validity-aware entropy concept that measures diversity only within the valid subspace specified by the reward function. Unlike standard entropy, which quantifies distributional spread over all trajectories, effective entropy focuses on the coverage of correct reasoning paths and ignores the internal distribution of incorrect outputs. This directly reframes exploration in RLVR: the central question is not how to increase uncertainty in general, but how to allocate uncertainty to valid reasoning.

To verify this idea, we instantiate a simple optimization method named \textbf{V}alidity-aware \textbf{E}ntropy \textbf{P}olicy \textbf{O}ptimization (VEPO). VEPO directly adds effective entropy as an auxiliary objective to GRPO. Its gradient induces an adaptive attraction dynamic: under-explored valid trajectories receive larger positive learning pressure, encouraging the policy to expand the support of correct solutions. Although the mathematical form of effective entropy contains a theoretical boundary at $\pi(o)>e^{-1}$, this regime is practically unreachable for non-trivial autoregressive LLM trajectories, whose sequence probabilities are products of many token probabilities. Therefore, in realistic LLM training, VEPO behaves as a pure attraction mechanism toward under-explored valid trajectories.

In this paper, our contributions are summarized as follows:
\begin{itemize}[leftmargin=1.2em]
\item We identify a critical structural misalignment in existing RLVR exploration, revealing that the reliance on standard entropy fosters ``the illusion of diversity'' by conflating hallucination with meaningful exploration. To address this, we establish \textit{Effective Entropy} as a validity-aware metric, and demonstrate that it serves as a superior optimization objective for reasoning alignment.
\item We instantiate VEPO, a deliberately simple method that directly incorporates effective entropy into RLVR training via an adaptive attraction mechanism, serving as a clean empirical validation of the effective entropy principle rather than a heavily engineered algorithm.
\item Experiments on six mathematical reasoning benchmarks across 1.5B and 7B LLMs show consistent aggregate accuracy gains over GRPO. Matched-count strategy measurements and paired training seeds substantiate useful diversity and repeatability within the tested setting.
\end{itemize}

\section{Related Work}
\label{sec:related_works}

\subsection{RLVR in LLMs}

RLVR has emerged as a distinct paradigm from Reinforcement Learning from Human Feedback (RLHF), anchoring LLM reasoning in objective correctness rather than subjective human ranking. Early approaches often relied on dense process supervision or value networks, such as PPO~\citep{schulman2017proximal}, but their computational burden limits scalability. Outcome-supervised RLVR has therefore gained increasing attention. GRPO~\citep{shao2024deepseekmath} is representative of this direction, estimating advantages via group statistics without an explicit value model. However, relying solely on outcome signals can lead to sparse feedback and premature convergence, making effective exploration design critical for further improving reasoning ability.

Rollout allocation provides a complementary route to improving RLVR learning signals. DynaMO~\citep{fang2026dynamo} allocates rollouts using gradient-variance information and modulates token advantages to address gradient attenuation. VEPO retains group-based sampling and instead changes which trajectory uncertainty receives an auxiliary incentive.

\subsection{Entropy-Aware Optimization}

Policy entropy is a standard metric for characterizing exploration diversity~\citep{williams1991function, eysenbach2021maximum}. In GRPO-style training, entropy often decays as the model converges, which may cause the policy to collapse onto limited reasoning patterns. Early methods mitigate this issue by adding entropy regularization~\citep{o2016combining} or modifying clipping rules. Recent methods include dynamic clipping strategies such as Clip-Higher~\citep{yu2025dapo}, Clip-Cov~\citep{cui2025entropy}, DCPO~\citep{yang2025dcpo}, and BAPO~\citep{xi2025bapo}, as well as gradient-preserving methods such as CISPO~\citep{chen2025minimax} and GPPO~\citep{su2025klear, su2025gppo}.

Soft trust-region methods further refine update geometry: DGPO~\citep{fu2026dgpo} applies bilateral decoupled decay to probability-gradient weights, while MASPO~\citep{fu2026maspo} combines soft gating with probability-mass adaptation and asymmetric signal reliability. These mechanisms regulate gradient utilization and stability; effective entropy specifies a validity-aware auxiliary objective.

More recent work shifts from blindly increasing entropy to controlling or utilizing it. AEPO~\citep{wang2025arbitrary} stabilizes the learning process by anchoring policy entropy to a dynamic target. ANCHOR~\citep{qin2026anchor} likewise studies stable entropy dynamics for efficient LLM reasoning, complementing our focus on the validity of the trajectories contributing to entropy. Entropy Advantage~\citep{cheng2025reasoning} and GTPO~\citep{tan2025gtpo} use entropy to modulate token-level advantages. Advantage Reweighting~\citep{yang2025not} down-weights overly uncertain tokens to focus learning on more reliable signals. Other works, such as HAPO~\citep{liu2025uniform} and ASPO~\citep{wang2025aspo}, further refine heterogeneous optimization behavior. Orthogonal theoretical perspectives also connect SFT and RL~\citep{qin2025supervised, wu2025generalization} or relate policy optimization to flow matching~\citep{zhu2025flowrl}.

Despite these advances, existing entropy-aware methods still treat entropy largely as a global uncertainty quantity. Our work focuses on a different question: whether the entropy being measured actually corresponds to valid reasoning diversity. Effective entropy answers this question by changing the domain of entropy measurement from the entire output space to the reward-verified valid subspace.

Adaptive sample weighting itself has a long history, including fuzzy memberships~\citep{lin2002fuzzy}, manifold regularization~\citep{belkin2006manifold}, and curricula~\citep{bengio2009curriculum}. Our contribution is the verifier-masked policy functional and its derived coefficient, rather than per-sample weighting in general. Effective mutual information~\citep{oh2025emi} instead adjusts query--response dependence against a reference joint distribution for distribution-shift analysis; it has a different domain and role from our fixed-query, verifier-gated output-entropy objective.

\section{Preliminary}
\label{sec:preliminary}

\paragraph{Problem Definition.}
We formulate the RLVR framework where a policy $\pi_\theta$ generates a reasoning trajectory $o$ conditioned on a query $q$ from dataset $Q$. The validity of each output is determined by a binary reward function $r(q,o)\in\{0,1\}$. In practice, the reward can be obtained from deterministic verifiers, such as exact-answer matching, mathematical verification, or compiler execution.

GRPO samples a group of $G$ outputs $\{o_i\}_{i=1}^G$ from the old policy $\pi_{\theta_{\mathrm{old}}}$ for each query and computes their rewards $R_i=r(q,o_i)$. The advantage of the $i$-th output is estimated by standardizing rewards within the group:
\begin{equation}
\hat{A}_i =
\frac{r(q,o_i)-\mathrm{mean}(\{R_i\}_{i=1}^G)}
{\mathrm{std}(\{R_i\}_{i=1}^G)}.
\end{equation}

GRPO then updates the policy by maximizing a clipped surrogate objective. Omitting the KL constraint for brevity, the objective is:
\begin{align}
\mathcal{J}_{\mathrm{GRPO}}(\theta)
=
\mathbb{E}_{q,\{o_i\}}
\frac{1}{G}\sum_{i=1}^G
\frac{1}{|o_i|}\sum_{t=1}^{|o_i|}
\min\Big[
\rho_{i,t}(\theta)\hat{A}_{i,t},
\mathrm{clip}(\rho_{i,t}(\theta),1-\epsilon,1+\epsilon)\hat{A}_{i,t}
\Big],
\end{align}
where
\begin{equation}
\rho_{i,t}(\theta)=
\frac{\pi_\theta(x_{t,i}\mid x_{<t,i},q)}
{\pi_{\theta_{\mathrm{old}}}(x_{t,i}\mid x_{<t,i},q)}
\end{equation}
is the token-level probability ratio.

\paragraph{Existing Use of Entropy.}
Entropy is commonly introduced to prevent premature convergence. A standard entropy-regularized objective adds an entropy term to the policy loss, while more recent methods use entropy to adapt clipping, rescale advantages, or control learning dynamics. These methods differ in implementation, but they share a common assumption: higher or better-controlled entropy during training implies better and more sustained exploration.

This assumption is incomplete in RLVR. Reasoning tasks care not only about distributional diversity, but also about whether the explored trajectories are valid. Standard entropy can increase even when the model spreads probability among invalid outputs. This motivates the need for a validity-aware entropy concept rather than blind entropy maximization.

\section{Effective Entropy}
\label{sec:effective_entropy}

\subsection{The Misalignment of Existing Entropy Metrics}
\label{sec:misalignment}

To motivate effective entropy, we first examine why common entropy metrics are structurally misaligned with reasoning validity in RLVR reasoning tasks.

\paragraph{Token Entropy Is Locally Blind.}
Token entropy is typically defined as
\begin{equation}
\mathcal{H}_{\mathrm{tok}}
=
\frac{1}{T}\sum_t
\mathcal{H}\big(\pi_\theta(\cdot\mid x_{<t},q)\big).
\end{equation}
It measures local uncertainty over the next-token distribution. However, in reasoning tasks, correctness is a property of the complete trajectory rather than an isolated token. A model can maintain high token entropy by randomizing locally plausible tokens that ultimately lead to invalid reasoning. Thus, token entropy is not a faithful proxy for useful reasoning exploration.

\paragraph{Sequence Entropy Is Globally Diluted.}
Sequence entropy is defined over the full trajectory space:
\begin{equation}
\mathcal{H}_{\mathrm{seq}}
=
-\sum_{o\in\mathcal{O}}
\pi_\theta(o\mid q)\log \pi_\theta(o\mid q).
\end{equation}
Although it captures global trajectory diversity, it treats all trajectories symmetrically. In reasoning tasks, the valid set is sparse while the invalid set is combinatorially dominant. Therefore, maximizing sequence entropy over $\mathcal{O}$ tends to allocate uncertainty to invalid outputs. A high value of $\mathcal{H}_{\mathrm{seq}}$ only indicates distributional spread, not valid reasoning diversity.

\subsection{Definition of Effective Entropy}

We introduce \textit{Effective Entropy} to measure the diversity of valid reasoning trajectories. Let $r(o\mid q)\in\{0,1\}$ be the binary correctness indicator. We define:
\begin{equation}
\label{eq:eff_entropy_def}
\mathcal{H}_{\mathrm{eff}}(\pi_\theta)
=
-\sum_{o\in\mathcal{O}}
r(o\mid q)\pi_\theta(o\mid q)\log \pi_\theta(o\mid q).
\end{equation}

This definition changes the domain of entropy measurement. Standard sequence entropy measures uncertainty over all trajectories, while effective entropy measures uncertainty weighted by validity. Invalid trajectories contribute zero, and only correct trajectories shape the metric. Effective entropy characterizes validity-weighted spread while also depending on valid probability mass; conditional strategy diversity requires a separate evaluation.

For model training, we express $\mathcal{H}_{\text{eff}}$ in its expectation form:
\begin{equation}
\label{eq:eff_entropy_exp}
\mathcal{H}_{\mathrm{eff}}(\pi_\theta)
=
\mathbb{E}_{o\sim\pi_\theta}
\left[-r(o\mid q)\log\pi_\theta(o\mid q)\right],
\end{equation}
which enables Monte Carlo estimation from sampled rollouts.

Effective entropy combines valid probability mass with conditional diversity, rather than measuring either alone (Appendix~\ref{app:valid_mass}). Terminal-verifier acceptance need not certify every reasoning step.

\subsection{Why Effective Entropy Changes the Exploration Paradigm}

The key role of effective entropy is conceptual rather than merely algorithmic. It suggests that exploration in RLVR should be evaluated by how much probability spread lies within the verified solution manifold. This view differs from conventional entropy maximization in two aspects.

First, it separates useful diversity from invalid dispersion. Standard entropy cannot distinguish a model that explores many correct strategies from one that produces many different wrong answers. Effective entropy directly encodes this distinction through the reward verifier.

Second, it provides a more aligned optimization signal. If a correct trajectory is under-explored, increasing its probability improves both reward learning and valid diversity. If an invalid trajectory is sampled, it receives no entropy-driven encouragement. This makes effective entropy naturally compatible with sparse verifiable rewards.

\section{VEPO: Instantiating Effective Entropy for RLVR}
\label{sec:methodology}

VEPO is a direct instantiation of effective entropy in RLVR training. It is intentionally simple: rather than designing a complex optimizer, we add effective entropy as an auxiliary objective to GRPO to verify whether validity-aware entropy provides a better exploration signal.

\begin{figure*}[t]
\centering
\includegraphics[width=1.0\linewidth]{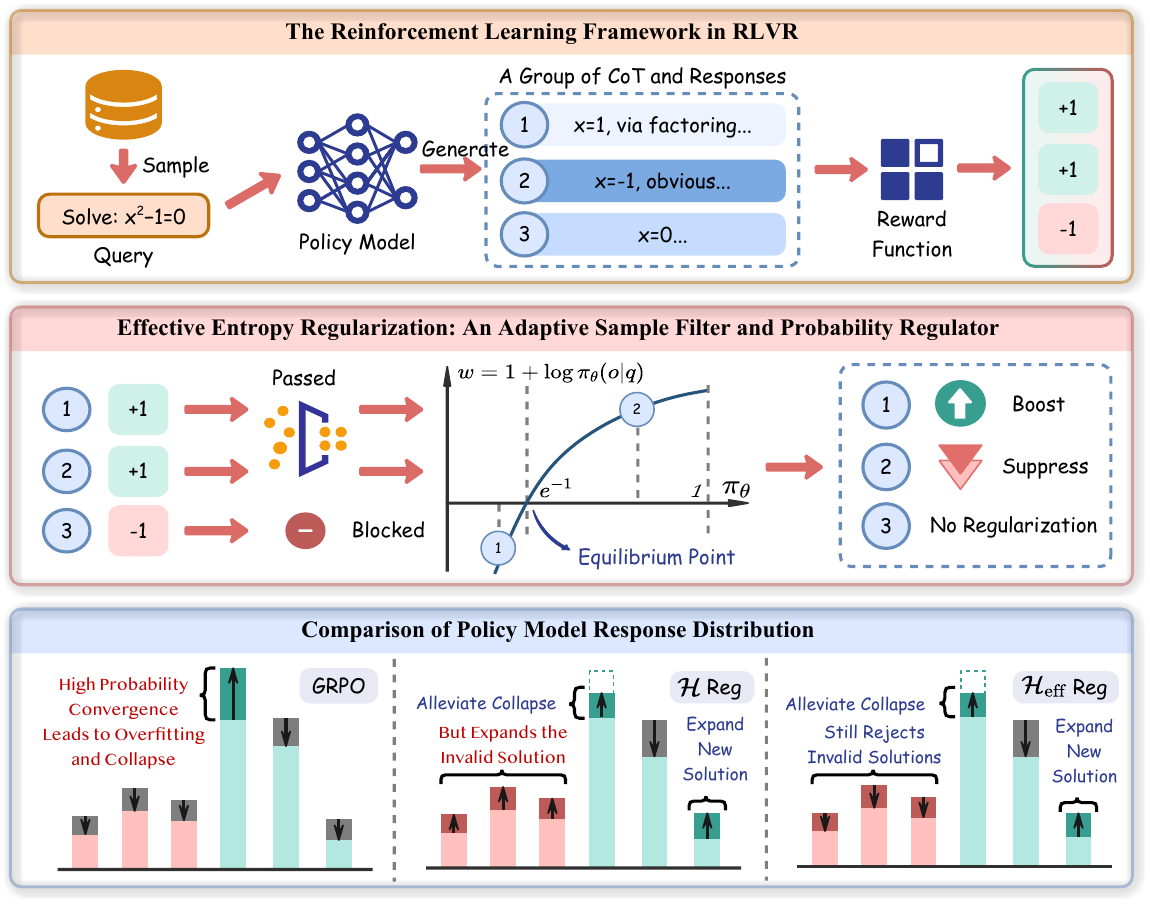}
\caption{Overview of VEPO. Effective entropy filters exploration through the reward-defined valid subspace. VEPO directly adds this signal to GRPO, inducing an adaptive attraction mechanism that increases the probability of under-explored valid trajectories while ignoring invalid ones.}
\vspace{-0.5em}
\label{fig:method}
\end{figure*}

\subsection{Objective}

We define the total optimization objective as:
\begin{equation}
\label{eq:total_loss}
\mathcal{J}_{\mathrm{VEPO}}(\theta)
=
\mathcal{J}_{\mathrm{GRPO}}(\theta)
+
\beta \cdot
\mathcal{H}_{\mathrm{eff}}(\pi_\theta),
\end{equation}
where $\beta>0$ controls the strength of effective-entropy regularization.

This objective keeps the original RLVR learning signal from GRPO and augments it with a validity-aware diversity signal. Since $\mathcal{H}_{\mathrm{eff}}$ ignores invalid trajectories, VEPO does not reward arbitrary uncertainty. Instead, it encourages the model to allocate additional probability mass to valid trajectories that are currently under-explored during policy optimization.

\subsection{Gradient Analysis}
\label{sec:gradient_analysis}

Let the effective entropy loss be
\begin{equation}
\mathcal{L}_{\mathrm{eff}}(\theta)
=
-\mathcal{H}_{\mathrm{eff}}(\pi_\theta).
\end{equation}
The gradient is:
\begin{equation}
\label{eq:eff_gradient}
\nabla_\theta \mathcal{L}_{\mathrm{eff}}(\theta)
=
\mathbb{E}_{o\sim\pi_\theta}
\left[
r(o\mid q)
\big(1+\log\pi_\theta(o\mid q)\big)
\nabla_\theta\log\pi_\theta(o\mid q)
\right].
\end{equation}

For valid trajectories, the scalar weight is:
\begin{equation}
\label{eq:weight}
w(o)=r(o\mid q)\big(1+\log\pi_\theta(o\mid q)\big).
\end{equation}
When $r(o\mid q)=1$ and $\pi_\theta(o\mid q)<e^{-1}$, we have $1+\log\pi_\theta(o\mid q)<0$. Minimizing the loss therefore increases the probability of this valid trajectory. In other words, VEPO attracts probability mass toward under-explored valid reasoning paths during RLVR training.

\paragraph{Theoretical Boundary and Practical LLM Regime.}
The mathematical form of Eq.~\ref{eq:eff_gradient} has a sign change when $\pi_\theta(o\mid q)>e^{-1}$. This would correspond to a boundary case where the objective decreases the probability of an overly dominant valid trajectory. However, this regime is practically unreachable for non-trivial autoregressive LLM outputs. A trajectory probability is the product of many conditional token probabilities; even moderately long reasoning chains have extremely small sequence probabilities. Therefore, in realistic LLM RLVR training, valid trajectories almost always satisfy $\pi_\theta(o\mid q)<e^{-1}$, and VEPO behaves as an adaptive attraction mechanism. We only mention this boundary for theoretical completeness and do not treat it as a practical training behavior.

\paragraph{Invalid Trajectories.}
For invalid trajectories, $r(o\mid q)=0$, so their effective-entropy weight is zero. They receive no entropy-driven attraction. This is the core distinction between effective entropy and standard entropy: VEPO does not spend the exploration budget on invalid outputs.

\subsection{Token-Level Implementation}

For autoregressive models, we decompose the sequence-level update into token-level gradients. For a sampled trajectory $o_i$, we compute the sequence-level weight:
\begin{equation}
\label{eq:seq_weight}
W_i =
r(o_i\mid q)
\left(1+\log\pi_\theta(o_i\mid q)\right).
\end{equation}
We detach this weight from the computation graph and apply it to the token log-probabilities:
\begin{equation}
\label{eq:token_gradient}
\nabla_\theta \mathcal{L}_{\mathrm{eff}}
\approx
\frac{1}{G}\sum_{i=1}^G
\sum_{t=1}^{T_i}
\mathrm{sg}[W_i]\cdot
\nabla_\theta\log\pi_\theta(x_{t,i}\mid x_{<t,i},q),
\end{equation}
where $\mathrm{sg}[\cdot]$ denotes stop-gradient.

This implementation is deliberately minimal. It does not introduce additional value networks, process reward models, or complex control mechanisms. Its purpose is to provide a clean test of whether effective entropy is a useful objective for RLVR exploration.

The sequence log-probability and token-gradient aggregation are both raw sums; response-length normalization would define a different functional. Sharing $W_i$ does not identify causally important reasoning tokens. Eq.~\ref{eq:eff_gradient} is exact for on-policy samples; reusing old-policy rollouts produces a local surrogate rather than an unbiased current-policy estimator. Appendix~\ref{app:loss_pseudocode} gives the corresponding detached-weight implementation, and Appendix~\ref{app:weight_controls} isolates correctness gating and length normalization.

\section{Experiments}
\label{sec:experiments}

\subsection{Experimental Setup}
\label{sec:experimental_setup}

\paragraph{Training Configuration.}
We evaluate VEPO on the DeepSeek-R1-Distill-Qwen family~\citep{guo2025deepseek}, using 1.5B and 7B backbones. All models are fine-tuned on DAPO-Math-17K~\citep{yu2025dapo}, a high-quality dataset for mathematical reasoning. Our training infrastructure follows the GRPO framework. We use a global batch size of 512 and a mini-batch size of 32. The group size is $G=8$ and the clipping threshold is fixed at $\epsilon=0.2$. The effective entropy coefficient $\beta$ is selected through grid search on a held-out validation set.

\paragraph{Verification Protocol.}
To ensure reliable RL signals, we use a strict dual-oracle verification protocol. During training, we use \texttt{math\_verify}~\citep{Kydlicek_Math-Verify_Math_Verification} for efficient online reward calculation. During evaluation, we use the more rigorous \texttt{prime\_math}~\citep{lightman2023let} to reduce reward hacking and improve final metric validity.

\paragraph{Evaluation Benchmarks and Metrics.}
We evaluate on AIME 2024~\citep{AIME24}, AIME 2025~\citep{AIME25}, AMC 2023~\citep{AMC}, MATH500~\citep{hendrycks2021measuring}, Minerva~\citep{lewkowycz2022solving}, and OlympiadBench~\citep{he2024olympiadbench}. We report two metrics. \textbf{Avg@32} measures expected accuracy over 32 sampled trajectories. \textbf{Pass@32} measures success under 32 repeated samples. It is affected by per-sample accuracy and does not directly measure semantic strategy diversity; we evaluate that property separately in Section~\ref{sec:direct_diversity}.

\paragraph{Baselines.}
We compare VEPO with vanilla GRPO and several entropy-aware baselines: \textbf{Entropy-Reg}~\citep{o2016combining}, which adds a standard entropy regularizer; \textbf{AEPO}~\citep{wang2025arbitrary}, which controls entropy around a dynamic target; \textbf{Entropy-Adv}~\citep{cheng2025reasoning} and \textbf{GTPO}~\citep{tan2025gtpo}, which use entropy to adjust learning signals; and \textbf{Adv-Reweighting}~\citep{yang2025not}, which down-weights certain uncertain or high-risk updates.

\subsection{Main Results}
\label{sec:main_results}

\begin{table*}[t]
\small
\centering
\setlength{\tabcolsep}{0.4pt}
\renewcommand{\arraystretch}{1.15}
\caption{Main results on mathematical reasoning benchmarks. We report Avg@32 (A@32) and Pass@32 (P@32). Bold indicates the best performance within each backbone.}
\label{tab:main_results}
\begin{tabular}{lcccccccccccccc}
\toprule
& \multicolumn{2}{c}{\textbf{AIME24}} & \multicolumn{2}{c}{\textbf{AIME25}} & \multicolumn{2}{c}{\textbf{AMC23}} & \multicolumn{2}{c}{\textbf{MATH500}} & \multicolumn{2}{c}{\textbf{Minerva}} & \multicolumn{2}{c}{\textbf{Olympiad}} & \multicolumn{2}{c}{\textbf{Average}} \\
\cmidrule(lr){2-3} \cmidrule(lr){4-5} \cmidrule(lr){6-7} \cmidrule(lr){8-9} \cmidrule(lr){10-11} \cmidrule(lr){12-13} \cmidrule(lr){14-15}
\textbf{Method} & A@32 & P@32 & A@32 & P@32 & A@32 & P@32 & A@32 & P@32 & A@32 & P@32 & A@32 & P@32 & A@32 & P@32 \\
\midrule
\multicolumn{15}{c}{\textit{Backbone: DeepSeek-R1-Distill-Qwen-1.5B}} \\
\midrule
GRPO & 33.2 & 71.8 & 27.7 & 49.9 & 79.5 & 94.8 & 77.6 & 90.8 & 26.1 & 48.8 & 46.3 & 64.7 & 48.4 & 70.1 \\
Entropy-Reg & 32.4 & 63.3 & \textbf{28.4} & 56.2 & 81.3 & 95.0 & 78.4 & 90.5 & 25.2 & 47.0 & 46.3 & 65.0 & 48.7 & 69.5 \\
Entropy-Adv & 29.6 & 67.8 & 23.9 & 44.2 & 77.1 & 94.1 & 78.5 & 89.7 & 25.2 & 48.6 & 46.4 & 64.3 & 46.8 & 68.1 \\
Adv-Reweighting & 25.0 & 59.1 & 22.3 & 40.7 & 70.6 & 94.6 & 75.6 & 90.1 & 25.1 & 51.1 & 42.3 & 63.2 & 43.5 & 66.5 \\
GTPO & 23.0 & 63.2 & 19.6 & 45.1 & 67.4 & 91.6 & 74.3 & 90.5 & 20.8 & 48.0 & 41.5 & 64.3 & 41.1 & 67.1 \\
AEPO & 33.9 & 64.9 & 28.1 & \textbf{57.1} & 80.2 & \textbf{97.2} & 77.4 & 91.3 & 24.2 & 47.1 & 46.9 & \textbf{66.5} & 48.5 & 70.7 \\
\textbf{VEPO} & \textbf{35.9} & \textbf{72.3} & 27.8 & 52.1 & \textbf{85.4} & 96.5 & \textbf{80.2} & \textbf{91.7} & \textbf{26.5} & \textbf{52.8} & \textbf{47.7} & 64.0 & \textbf{50.6} & \textbf{71.6} \\
\midrule
\multicolumn{15}{c}{\textit{Backbone: DeepSeek-R1-Distill-Qwen-7B}} \\
\midrule
GRPO & 48.2 & 82.5 & 37.4 & 60.5 & 88.1 & 96.6 & 84.8 & 92.4 & 37.4 & 57.2 & 57.2 & 73.9 & 58.9 & 77.2 \\
Entropy-Reg & 49.7 & 82.7 & 36.1 & 60.5 & 89.4 & 97.9 & 84.9 & 92.3 & 37.5 & 54.6 & 57.9 & 73.5 & 59.2 & 76.9 \\
Entropy-Adv & 48.1 & 83.5 & 36.8 & 61.5 & 88.1 & \textbf{99.8} & 84.1 & 91.3 & 37.4 & 56.1 & 56.6 & 73.5 & 58.5 & 77.6 \\
Adv-Reweighting & 43.0 & 76.3 & 31.9 & 56.4 & 83.8 & 94.4 & 83.7 & 93.9 & 35.4 & 56.5 & 55.9 & 74.7 & 55.6 & 75.4 \\
GTPO & 43.8 & 83.4 & 32.0 & \textbf{63.5} & 85.1 & 95.0 & 83.8 & \textbf{95.3} & 32.8 & 57.6 & 54.5 & 73.6 & 55.3 & 78.1 \\
AEPO & 50.0 & 78.1 & 37.9 & 61.9 & 87.9 & 97.4 & 85.4 & 95.1 & 36.6 & \textbf{59.4} & 58.1 & \textbf{76.7} & 59.3 & 78.1 \\
\textbf{VEPO} & \textbf{51.3} & \textbf{86.0} & \textbf{41.1} & 63.4 & \textbf{91.5} & 97.5 & \textbf{85.5} & 92.6 & \textbf{38.3} & 57.4 & \textbf{59.0} & 74.2 & \textbf{61.1} & \textbf{78.5} \\
\bottomrule
\end{tabular}
\vspace{-0.5em}
\end{table*}

Table~\ref{tab:main_results} summarizes the main results. VEPO improves the six-benchmark aggregate over vanilla GRPO across both 1.5B and 7B backbones, while individual Pass@32 comparisons are mixed. On the 1.5B model, VEPO achieves the best average Avg@32 and Pass@32 among the compared methods. On the 7B model, VEPO also obtains the strongest average Avg@32 and a competitive Pass@32. These results suggest that effective entropy provides a more aligned exploration signal than standard entropy or entropy-derived advantage reshaping methods.

The key point is not that VEPO is a sophisticated optimizer, but that even this simple instantiation of effective entropy yields consistent gains. This supports our central claim: measuring diversity within the valid subspace provides a better foundation for RLVR exploration than measuring entropy over all tokens or all sampled trajectories.

\begin{figure*}[t]
\centering
\begin{subfigure}{0.36\textwidth}
\centering
\includegraphics[width=\linewidth]{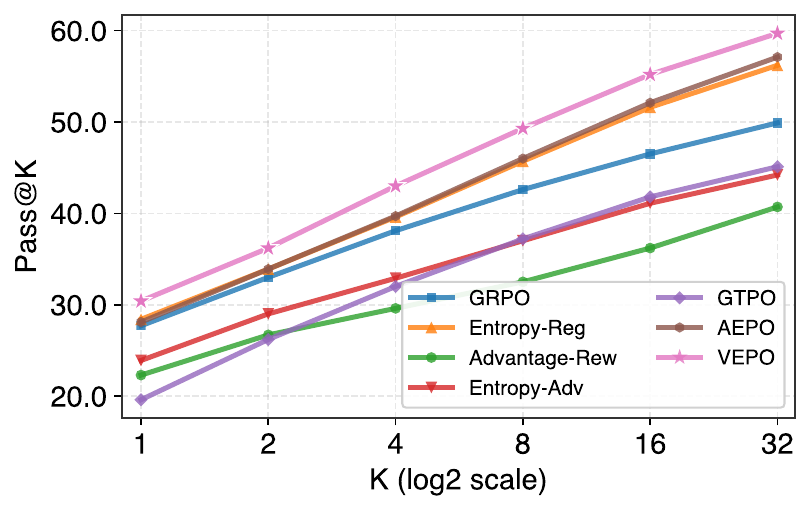}
\caption{Pass@K, 1.5B}
\label{fig:main_1}
\end{subfigure}
\hfill
\begin{subfigure}{0.31\textwidth}
\centering
\includegraphics[width=\linewidth]{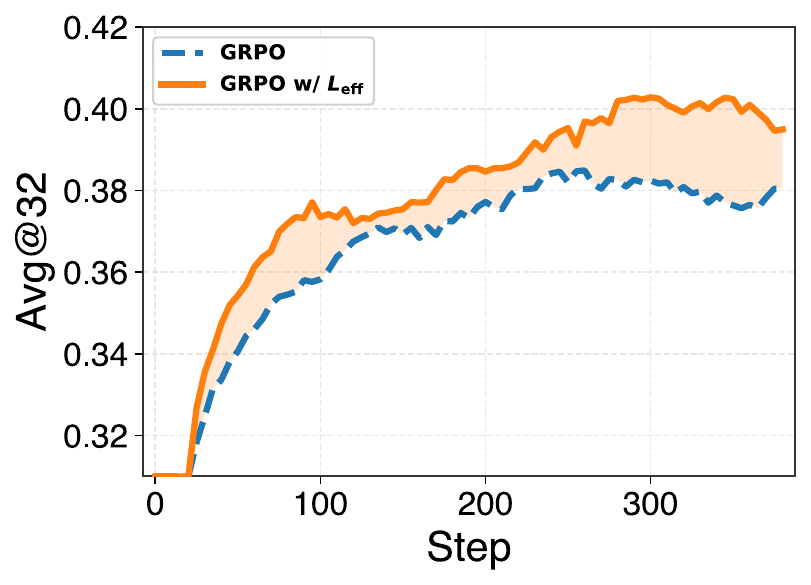}
\caption{Avg@32}
\label{fig:main_2}
\end{subfigure}
\hfill
\begin{subfigure}{0.31\textwidth}
\centering
\includegraphics[width=\linewidth]{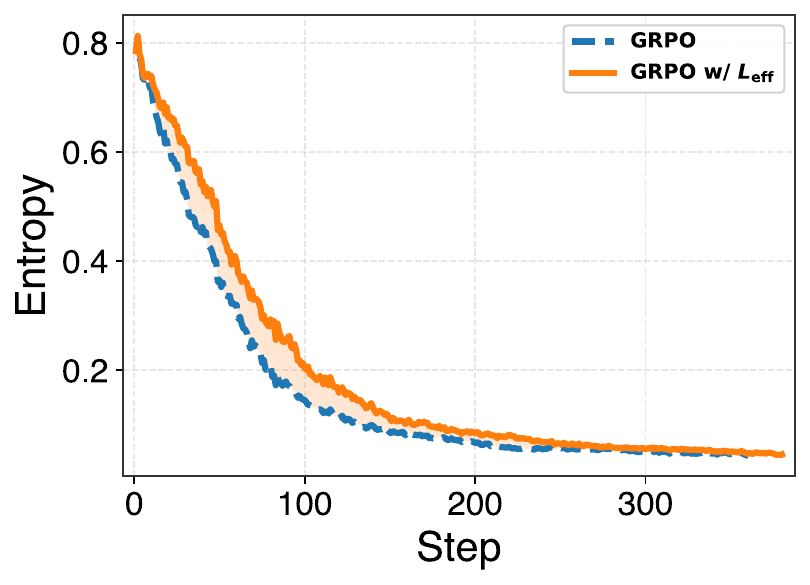}
\caption{Token entropy}
\label{fig:main_3}
\end{subfigure}

\vspace{0.45em}

\begin{subfigure}{0.36\textwidth}
\centering
\includegraphics[width=\linewidth]{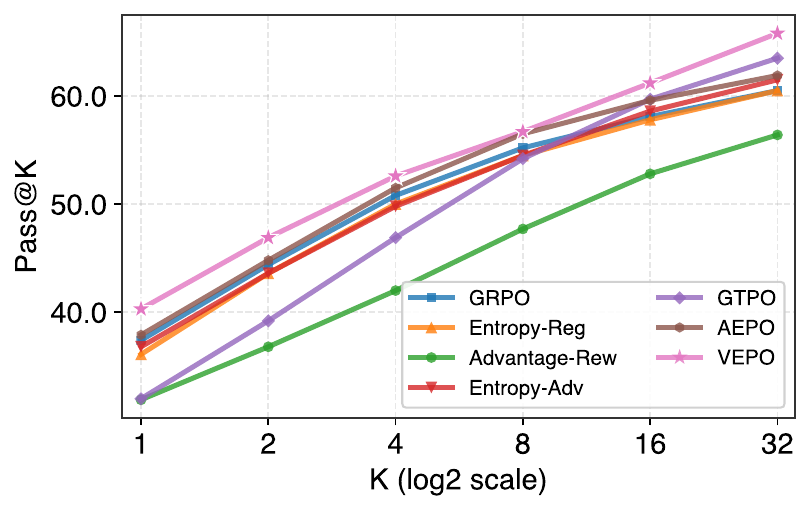}
\caption{Pass@K, 7B}
\label{fig:main_4}
\end{subfigure}
\hfill
\begin{subfigure}{0.31\textwidth}
\centering
\includegraphics[width=\linewidth]{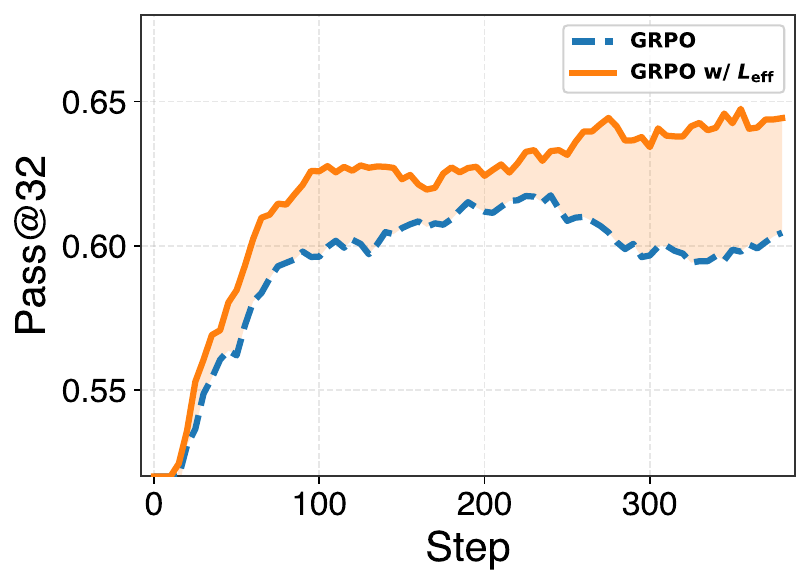}
\caption{Pass@32}
\label{fig:main_5}
\end{subfigure}
\hfill
\begin{subfigure}{0.31\textwidth}
\centering
\includegraphics[width=\linewidth]{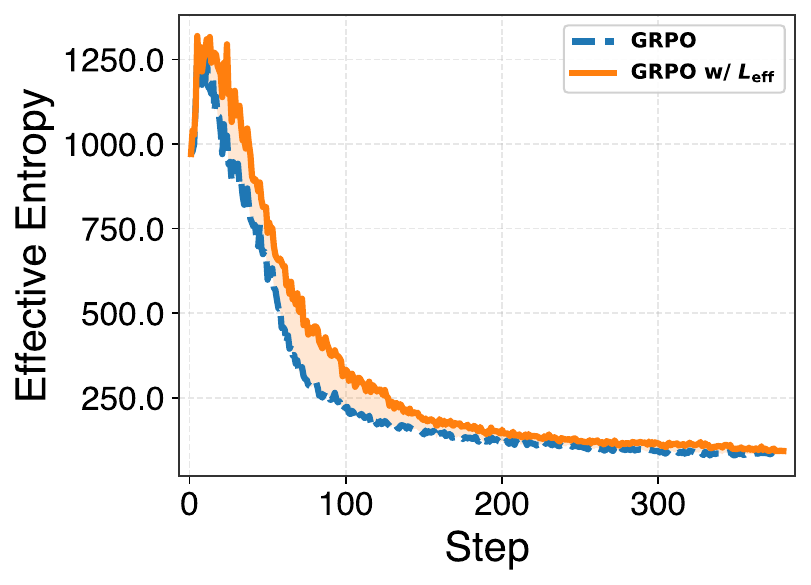}
\caption{Effective entropy}
\label{fig:main_6}
\end{subfigure}

\caption{Main empirical results. Subfigures (a) and (d) show Pass@K on AIME 2025 for the 1.5B and 7B models. Subfigures (b) and (e) show Avg@32 and Pass@32 training dynamics on the 7B model. Subfigures (c) and (f) compare token entropy and effective entropy during training. Together, these results show that VEPO improves performance while sustaining stronger validity-aware exploration across different model scales.}
\vspace{-0.5em}
\label{fig:main_results}
\end{figure*}

\paragraph{Repeated-Sampling Success and Training Dynamics.}
Figure~\ref{fig:main_results}(a,d) reports repeated-sampling success across $k$; these curves do not independently establish semantic strategy diversity. Panels (b,e) show AIME 2025 training dynamics, while (c,f) track token and effective entropy. VEPO sustains learning and higher effective entropy in the tested setting.

\subsection{Direct Strategy Diversity and Statistical Reliability}
\label{sec:direct_diversity}

To separate strategy diversity from accuracy, we compare GRPO and VEPO on the 441 MATH500-7B problems where both produce at least eight correct trajectories in their 32-sample pools. We match exactly eight correct trajectories per method per problem, average over 10,000 rarefactions, and cluster answer-stripped reasoning using frozen gte-Qwen2 embeddings~\citep{gteqwen2024} and a method-blinded, validated threshold. Appendix~\ref{app:diversity_checks} details the protocol and robustness checks.

We strip final-answer spans, pool the remaining chains of thought from both methods, and cluster within each problem using frozen gte-Qwen2-7B-instruct embeddings~\citep{gteqwen2024}, L2 normalization, and average-linkage agglomerative clustering with cosine distance~\citep{mullner2011modern}. A strategy denotes a shared mathematical plan, invariant, decomposition, or substitution, irrespective of wording or notation. Before comparing methods, we fix the pipeline and select one global distance threshold on method-blinded development annotations. Two mathematical annotators label 600 same/different-strategy pairs with Cohen's $\kappa=0.82$; the frozen pipeline achieves precision 0.88 and recall 0.84 on a disjoint held-out set.

\begin{table}[t]
\centering
\small
\caption{Direct strategy diversity on the 441 commonly eligible MATH500-7B problems. Each method contributes exactly eight correct trajectories per problem. Hill-1 is the effective number of strategy modes~\citep{hill1973diversity}.}
\label{tab:direct_diversity}
\begin{tabular}{lccc}
\toprule
Metric & GRPO & VEPO & Difference \\
\midrule
Distinct clusters/problem & 2.31 & \textbf{2.68} & $+0.37$ \\
Hill-1 effective modes & 1.97 & \textbf{2.24} & $+0.27$ \\
Largest-cluster share & 0.68 & \textbf{0.61} & $-0.07$ \\
\bottomrule
\end{tabular}
\end{table}

The paired cluster-count gain is $+0.37$ (16.0\%), with 95\% problem-bootstrap CI $[0.24,0.50]$ and sign-flip permutation $p<10^{-4}$. Matching correct counts prevents this increase from arising mechanically from higher accuracy. This finite-sample endpoint concerns commonly solved, answer-verified prompts; it does not certify every reasoning step.

Three paired training seeds ($0,1,2$) yield six-benchmark Avg@32 gains of $+2.1,+2.4,+2.1$ percentage points at 1.5B and $+2.3,+2.4,+1.9$ at 7B. All paired Avg@32 and Pass@32 gains are positive (Table~\ref{tab:paired_seeds}); three seeds support repeatability without strong seed-level significance claims.

\subsection{Hyperparameter Analysis}
\label{sec:hyperparam}

\begin{figure*}[t]
\centering
    \begin{subfigure}{0.24\columnwidth}
        \centering
        \includegraphics[width=\linewidth]{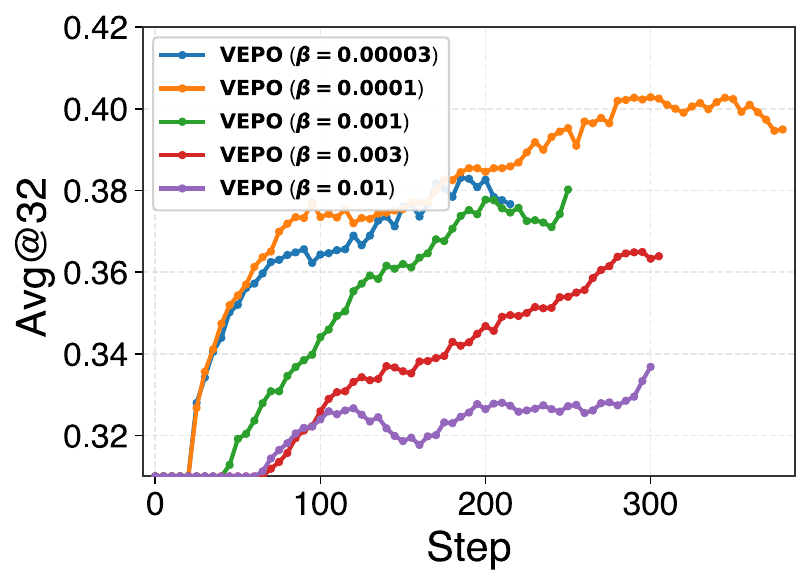}
        \caption{AIME25 Avg@32}
        \label{fig:dyn_hyper_1}
    \end{subfigure}
    \hfill
    \begin{subfigure}{0.24\columnwidth}
        \centering
        \includegraphics[width=\linewidth]{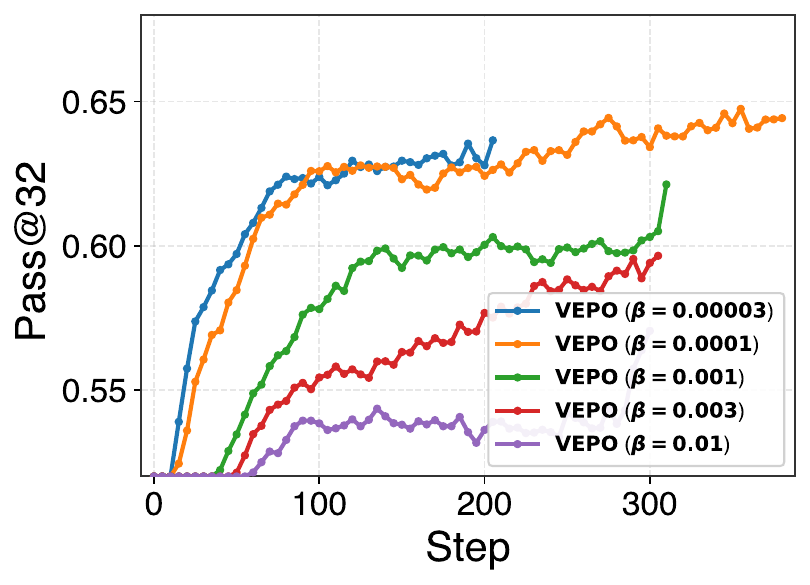}
        \caption{AIME25 Pass@32}
        \label{fig:dyn_hyper_2}
    \end{subfigure}
    \hfill
    \begin{subfigure}{0.24\columnwidth}
        \centering
        \includegraphics[width=\linewidth]{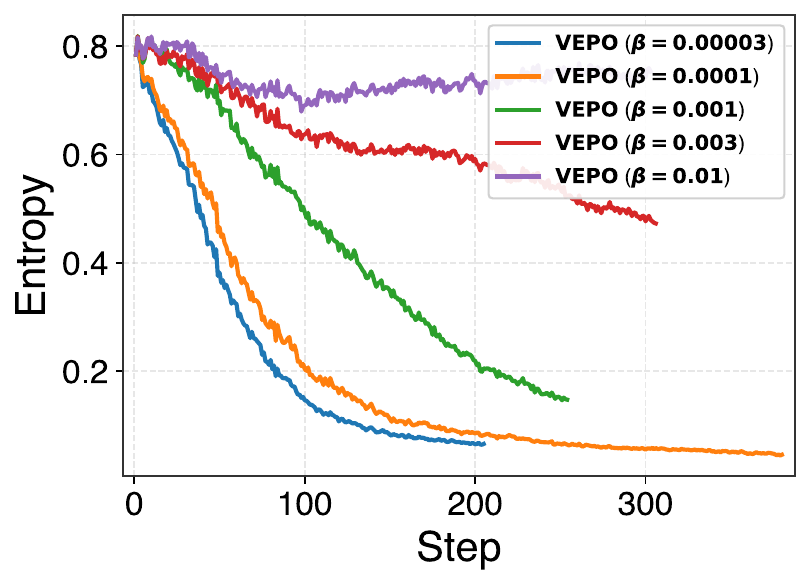}
        \caption{Token entropy}
        \label{fig:dyn_hyper_3}
    \end{subfigure}
    \hfill
    \begin{subfigure}{0.24\columnwidth}
        \centering
        \includegraphics[width=\linewidth]{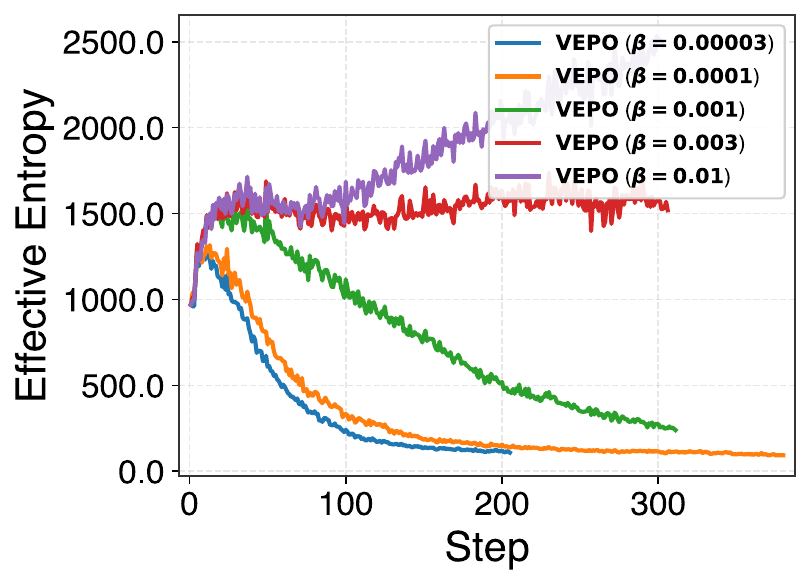}
        \caption{Effective entropy}
        \label{fig:dyn_hyper_4}
    \end{subfigure}
\caption{Impact of the effective entropy coefficient $\beta$. Larger $\beta$ maintains stronger exploration but can delay convergence. A moderate coefficient provides the best balance between valid diversity and optimization efficiency across training stages.}
\vspace{-0.5em}
\label{fig:hyperparam}
\end{figure*}

We conduct hyperparameter experiments with $\beta\in\{3\mathrm{e}{-5},1\mathrm{e}{-4},1\mathrm{e}{-3},3\mathrm{e}{-3},1\mathrm{e}{-2}\}$. As shown in Figure~\ref{fig:hyperparam}, effective entropy regularization affects the balance between exploration and convergence. Larger $\beta$ values maintain higher entropy and effective entropy, but may keep the model in an active exploration phase for too long. In contrast, a moderate value such as $\beta=1\mathrm{e}{-4}$ provides sufficient valid-space exploration while allowing efficient convergence. This is used as our default.

\subsection{Attribution Analysis}
\label{sec:analysis}

To understand why effective entropy improves RLVR training, we compare VEPO with GRPO and standard Entropy-Reg. We analyze three aspects: the magnitude of exploration, the structural quality of exploration, and the conversion efficiency from exploration to performance.

\begin{figure*}[t]
\centering
    \begin{subfigure}{0.24\columnwidth}
        \centering
        \includegraphics[width=\linewidth]{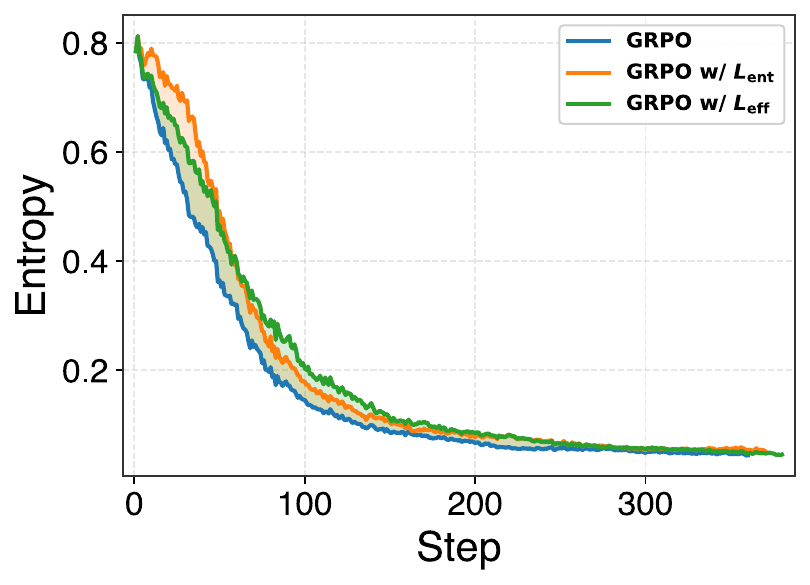}
        \caption{Token entropy}
        \label{fig:dynamics_entropy}
    \end{subfigure}
    \hfill
    \begin{subfigure}{0.24\columnwidth}
        \centering
        \includegraphics[width=\linewidth]{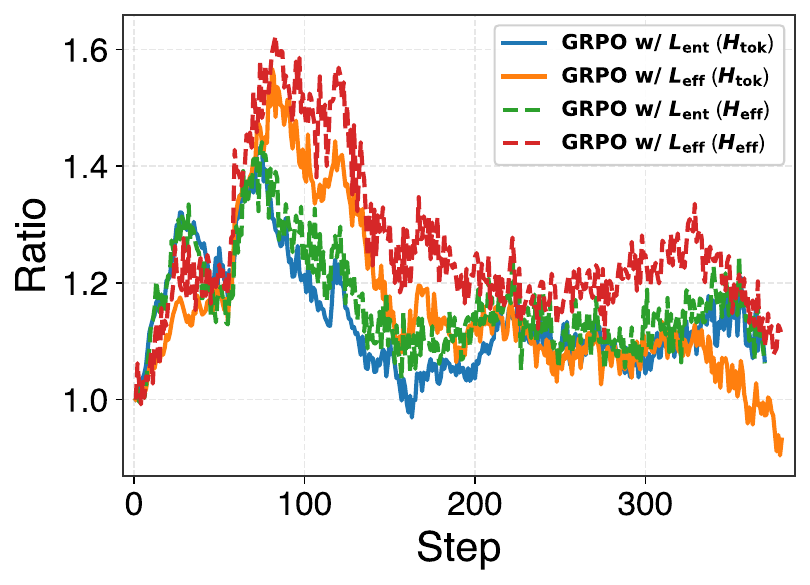}
        \caption{Relative entropy gain}
        \label{fig:dynamics_gain}
    \end{subfigure}
    \hfill
    \begin{subfigure}{0.24\columnwidth}
        \centering
        \includegraphics[width=\linewidth]{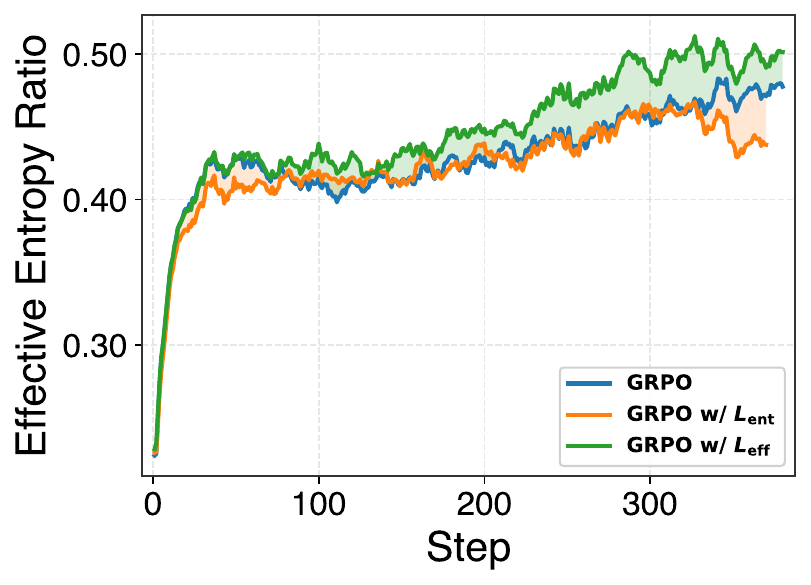}
        \caption{Effective ratio}
        \label{fig:dynamics_ratio}
    \end{subfigure}
    \hfill
    \begin{subfigure}{0.24\columnwidth}
        \centering
        \includegraphics[width=\linewidth]{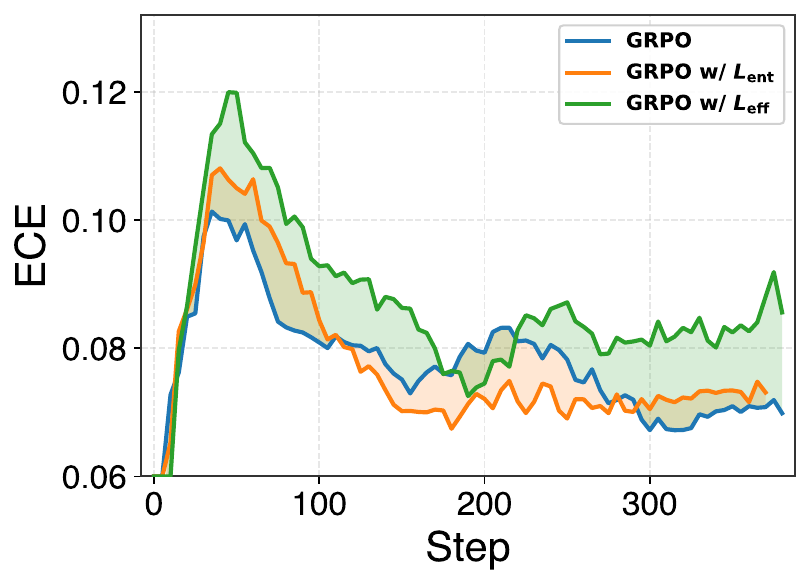}
        \caption{Conversion efficiency}
        \label{fig:dynamics_ece}
    \end{subfigure}
\caption{Attribution analysis. Entropy-Reg increases raw entropy but does not reliably improve the structural quality of exploration. VEPO allocates more exploration to the valid subspace, leading to higher effective ratio and stronger conversion efficiency.}
\vspace{-0.5em}
\label{fig:attribution}
\end{figure*}

\paragraph{Magnitude and Direction.}
Figure~\ref{fig:dynamics_entropy} shows that both VEPO and Entropy-Reg can maintain higher token entropy than GRPO. However, similar entropy magnitude does not imply similar exploration quality. To distinguish the direction of exploration more precisely, we define the relative entropy gain of a method $M$ as:
\begin{equation}
G(M)=\frac{\mathcal{H}_M}{\mathcal{H}_{\mathrm{GRPO}}}.
\end{equation}
As shown in Figure~\ref{fig:dynamics_gain}, Entropy-Reg increases total entropy and effective entropy in a coupled manner, suggesting a largely blind expansion of the distribution. In contrast, VEPO yields a stronger gain in effective entropy relative to total entropy. This indicates that VEPO preferentially expands the valid subspace rather than merely spreading probability over all outputs.

\paragraph{Structural Quality.}
We further define the effective ratio:
\begin{equation}
\rho_{\mathrm{eff}}
=
\frac{\mathcal{H}_{\mathrm{eff}}}{\mathcal{H}_{\mathrm{seq}}}.
\end{equation}
This metric measures the fraction of sequence-level uncertainty that is allocated to valid trajectories. As shown in Figure~\ref{fig:dynamics_ratio}, Entropy-Reg can reduce the effective ratio in later training stages, implying that blind entropy regularization may dilute the policy's focus on correctness. VEPO maintains a higher effective ratio, showing that its exploration is structurally cleaner and more consistently aligned with verified reasoning.

\paragraph{Conversion Efficiency.}
Finally, we evaluate whether additional exploration translates into performance gain. Let $\Delta\mathcal{P}$ be the performance gain and $\Delta\mathcal{H}$ be the entropy change relative to initialization. We define Entropy Conversion Efficiency (ECE) as:
\begin{equation}
\mathrm{ECE}
=
\Delta\mathcal{P}\cdot
\exp\left(
\mathrm{sgn}(\Delta\mathcal{P})
\cdot
\frac{\Delta\mathcal{H}}{\mathcal{H}_{\mathrm{init}}}
\right).
\end{equation}
As shown in Figure~\ref{fig:dynamics_ece}, Entropy-Reg provides only transient benefit and later converges toward the GRPO baseline. VEPO maintains a higher ECE trajectory, indicating that validity-aware exploration is more effectively converted into reasoning performance. These diagnostics describe allocation and conversion of entropy during optimization. The matched-correct-count analysis in Section~\ref{sec:direct_diversity} provides a complementary, direct assessment of strategy diversity while keeping the number of correct samples fixed within each problem.

\par\vfill
\section{Conclusion}

In this paper, we identify a structural weakness in conventional entropy-based exploration for RLVR. Standard entropy measures distributional spread without considering whether the explored trajectories are valid, thereby conflating useful reasoning diversity with invalid dispersion. To address this issue, we introduce \textit{Effective Entropy}, a validity-aware entropy concept that measures diversity only within the reward-verified valid subspace of reasoning trajectories.

Based on effective entropy, we instantiate VEPO, a simple and direct optimization method that adds effective entropy to GRPO. VEPO induces an adaptive attraction mechanism toward under-explored valid reasoning trajectories and ignores invalid ones. Experiments on six mathematical reasoning benchmarks show consistent aggregate accuracy gains over GRPO and competitive repeated-sampling success against entropy-aware baselines.

Our mathematical RLVR results support validity-aware exploration on the tested backbones, rather than universal transfer. Direct matched-count strategy measurements and paired training seeds strengthen the evidence beyond entropy and pass rates alone. Larger models, interactive agents, process-level verification, and open-ended rewards remain directions for future work.

\clearpage
\begin{ack}
\paragraph{Funding and Computing Support.}
Meituan provided the computing resources used in this work. The study was initiated by employees as independent research and was not part of a formally designated Meituan project. The authors received no other funding or support for this work.

\paragraph{Competing Interests.}
Xiaoliang Fu, Yangyi Fang, Cong Qin, Binbin Zheng, and Zekai Shao were previously employed by Meituan. Jiaye Lin and Chaowen Hu list Meituan as their current affiliation. The authors declare no other competing interests.
\end{ack}

\bibliographystyle{abbrvnat}
\bibliography{reference}

\newpage
\appendix

\section{Why Effective Entropy: Domain and Focus of Exploration}
\label{app:entropy_analysis}

In this section, we provide a comparative analysis of why Effective Entropy provides the correct ``focus'' for reasoning tasks compared to standard formulations. The core distinction lies not in the calculation method, but in the \textbf{domain} over which the uncertainty is measured.

\paragraph{Justification of the Inequality $|\mathcal{O}_{\text{invalid}}| \gg |\mathcal{O}_{\text{valid}}|$.}
The assumption that the invalid subspace combinatorially dominates the valid subspace is fundamental to our critique of standard entropy. We justify this formally through the lens of combinatorial expansion and logical fragility:
\begin{enumerate}
    \item \textbf{Total Space Expansion:} Consider a reasoning chain of length $L$ generated from a vocabulary $\mathcal{V}$. The total output space cardinality is $|\mathcal{O}| = |\mathcal{V}|^L$. For modern LLMs, $|\mathcal{V}| \approx 10^5$, creating a hyper-astronomical search space.
    \item \textbf{Logical Sparsity:} Validity is a \textit{strict constraint satisfaction problem}. A valid reasoning chain requires a precise sequence of logical transitions. If we model a reasoning path as a chain of $K$ logical steps, and at each step only a small subset of tokens $\mathcal{V}_{\text{correct}} \subset \mathcal{V}$ maintains logical consistency, the size of the valid manifold scales as $|\mathcal{O}_{\text{valid}}| \approx |\mathcal{V}_{\text{correct}}|^L$.
    \item \textbf{Exponential Divergence:} Since $|\mathcal{V}_{\text{correct}}| \ll |\mathcal{V}|$ (e.g., there are few correct numbers but infinite incorrect ones), the ratio of valid to total sequences is:
    \begin{equation}
        \frac{|\mathcal{O}_{\text{valid}}|}{|\mathcal{O}|} \approx \left( \frac{|\mathcal{V}_{\text{correct}}|}{|\mathcal{V}|} \right)^L \to 0 \quad \text{as } L \to \infty
    \end{equation}
    Thus, $\mathcal{O}_{\text{invalid}} = \mathcal{O} \setminus \mathcal{O}_{\text{valid}}$ constitutes the overwhelming majority of the configuration space. Any metric maximizing distribution spread over $\mathcal{O}$ without constraints (like $\mathcal{H}_{\text{seq}}$) will statistically force the probability mass into $\mathcal{O}_{\text{invalid}}$ due to this dimensional curse.
\end{enumerate}

\paragraph{Domain Comparison of Metrics.}
Let $\mathcal{O}$ be the set of all possible output sequences.

\begin{itemize}
    \item \textbf{Token Entropy ($\mathcal{H}_{\text{tok}}$)} measures uncertainty over the vocabulary $V$ at each step $t$.
    \begin{itemize}
        \item \textit{Focus:} Local next-token prediction.
        \item \textit{Limitation:} It is agnostic to the set $\mathcal{O}_{\text{valid}}$. A model can have high $\mathcal{H}_{\text{tok}}$ by randomizing tokens that lead to $o \notin \mathcal{O}_{\text{valid}}$.
    \end{itemize}
    
    \item \textbf{Sequence Entropy ($\mathcal{H}_{\text{seq}}$)} measures uncertainty over the full set $\mathcal{O}$.
    \begin{itemize}
        \item \textit{Focus:} Global distribution spread.
        \item \textit{Limitation:} It treats $\mathcal{O}_{\text{valid}}$ and $\mathcal{O}_{\text{invalid}}$ symmetrically. Given the inequality $|\mathcal{O}_{\text{invalid}}| \gg |\mathcal{O}_{\text{valid}}|$ established above, maximizing this metric predominantly focuses on the invalid subspace.
    \end{itemize}
    
    \item \textbf{Effective Entropy ($\mathcal{H}_{\text{eff}}$)} measures uncertainty weighted by the validity indicator.
    \begin{itemize}
        \item \textit{Focus:} Distribution spread strictly within $\mathcal{O}_{\text{valid}}$.
        \item \textit{Advantage:} It mathematically ignores the internal dynamics of invalid sequences. It rewards the model only for expanding the support of the distribution covering correct answers.
    \end{itemize}
\end{itemize}

\section{Derivations and Theoretical Analysis}
\label{app:proofs}

\subsection{Theoretical Analysis of Global Optima}
\label{app:optima_analysis}

To rigorously justify why effective entropy provides a more aligned exploration principle for RLVR, we analyze the global optima of three objectives: standard reward maximization, standard entropy regularization, and effective-entropy regularization.

Let $\mathcal{S} = \mathcal{O}_{\text{valid}}$ be the set of correct answers and $\mathcal{N} = \mathcal{O}_{\text{invalid}}$ be the set of incorrect answers, such that the output space $\mathcal{O} = \mathcal{S} \cup \mathcal{N}$. Let the reward function be binary: $r(o)=1$ if $o \in \mathcal{S}$ and $0$ otherwise. We assume $|\mathcal{N}| \gg |\mathcal{S}|$.

\paragraph{Case 1: Standard RL (Max Reward).}
The objective is $\max_\pi \mathbb{E}_{o \sim \pi}[r(o)] = \sum_{o \in \mathcal{S}} \pi(o)$.
\begin{itemize}
    \item \textbf{Solution Space:} The maximum possible reward is 1. Any policy $\pi$ such that $\sum_{o \in \mathcal{S}} \pi(o) = 1$ is a global optimum.
    \item \textit{Analysis:} The optimization landscape is degenerate. Without a diversity objective, a \textbf{Dirac distribution} on one solution $o^* \in \mathcal S$ is permitted, but the reward objective alone does not imply that every optimizer converges to such a distribution.
\end{itemize}

\paragraph{Case 2: Standard Entropy Regularization (The Illusion of Diversity).}
We consider the standard regularization objective where diversity is encouraged over the entire output space $\mathcal{O}$:
\begin{equation}
    \mathcal{J}_{\text{reg}}(\pi) = \mathbb{E}[r] + \beta \mathcal{H}_{\text{seq}} = \sum_{o \in \mathcal{O}} \pi(o)r(o) - \beta \sum_{o \in \mathcal{O}} \pi(o)\log\pi(o)
\end{equation}
Using Lagrange multipliers with constraint $\sum \pi(o) = 1$, the Lagrangian is:
\begin{equation}
    \mathcal{L} = \sum_{o \in \mathcal{O}} \pi(o)r(o) - \beta \sum_{o \in \mathcal{O}} \pi(o)\log\pi(o) + \gamma (\sum_{o \in \mathcal{O}} \pi(o) - 1)
\end{equation}
Solving $\frac{\partial \mathcal{L}}{\partial \pi(o)} = 0$ yields the standard Boltzmann (Gibbs) distribution:
\begin{equation}
    \pi^*(o) = \frac{\exp(r(o)/\beta)}{Z}, \quad \text{where } Z = \sum_{o' \in \mathcal{O}} \exp(r(o')/\beta)
\end{equation}
\begin{itemize}
    \item \textbf{Leakage into Invalid Subspace:}
    For any valid solution $o_s \in \mathcal{S}$ and invalid solution $o_n \in \mathcal{N}$:
    \begin{equation}
        \pi^*(o_s) \propto e^{1/\beta}, \quad \pi^*(o_n) \propto e^{0/\beta} = 1
    \end{equation}
    Crucially, $\pi^*(o_n) \neq 0$. The optimal policy \textit{must} assign non-zero probability to invalid answers to satisfy the entropy constraint.
    \item \textit{Analysis (The Illusion):} Although $\pi^*(o_s) > \pi^*(o_n)$, the total probability mass assigned to hallucinations is:
    \begin{equation}
        P(\text{Error}) = \frac{|\mathcal{N}|}{|\mathcal{N}| + |\mathcal{S}|e^{1/\beta}}
    \end{equation}
    Since $|\mathcal{N}|$ (combinatorial errors) is typically orders of magnitude larger than $|\mathcal{S}|$, even for moderate $\beta$, significant probability mass is forced into the invalid region. This mathematically proves that standard entropy regularization creates an "illusion of diversity" by diluting validity with inevitable hallucinations.
\end{itemize}

\paragraph{Case 3: VEPO (Max Reward + Effective Entropy).}
We consider the VEPO objective:
\begin{equation}
    \mathcal{J}_{\text{vepo}}(\pi) = \mathbb{E}[r] + \beta \mathcal{H}_{\text{eff}} = \sum_{o \in \mathcal{O}} \pi(o)r(o) - \beta \sum_{o \in \mathcal{O}} r(o)\pi(o)\log\pi(o)
\end{equation}
Substituting the binary reward structure ($r(o)=0$ for $o \in \mathcal{N}$ eliminates terms in both sums):
\begin{equation}
    \mathcal{J}_{\text{vepo}}(\pi) = \sum_{o \in \mathcal{S}} \pi(o) (1 - \beta \log \pi(o))
\end{equation}
\paragraph{Scope and Coefficient Condition.}
We analyze a finite trajectory simplex with a nonempty valid set and $0<\beta\le1$, which includes all tested coefficients. For a valid coordinate, the derivative of $\pi(o)[1-\beta\log\pi(o)]$ is $1-\beta[1+\log\pi(o)]$, positive whenever $\pi(o)<1$. Consequently, any positive invalid mass can be shifted to a valid coordinate to improve the objective. This condition is required for the argument below.

\textbf{Step 1: Probability Assignment to Invalid Sets.}
By the preceding derivative argument, the optimum under the stated coefficient condition has $\pi^*(o)=0$ for all $o\in\mathcal N$.

\textbf{Step 2: Distribution within Valid Sets.}
We solve for $\pi$ within $\mathcal{S}$ using Lagrange multipliers:
\begin{equation}
    \mathcal{L} = \sum_{i \in \mathcal{S}} \pi_i (1 - \beta \log \pi_i) + \gamma \left( \sum_{i \in \mathcal{S}} \pi_i - 1 \right)
\end{equation}
Setting $\frac{\partial \mathcal{L}}{\partial \pi_k} = 0$:
\begin{equation}
    1 - \beta(1 + \log \pi_k) + \gamma = 0 \implies \pi_k = \exp\left( \frac{1 - \beta + \gamma}{\beta} \right)
\end{equation}
Since the RHS is constant for all $k \in \mathcal{S}$, we have $\pi_i = \pi_j$.
\begin{itemize}
    \item \textbf{Conclusion:} The unique global optimum is the \textbf{Uniform Distribution} strictly over the set of correct answers: 
    \begin{equation}
        \pi^*(o) = \begin{cases} \frac{1}{|\mathcal{S}|} & \text{if } o \in \mathcal{S} \\ 0 & \text{if } o \in \mathcal{N} \end{cases}
    \end{equation}
    This is an optimum of the idealized expected-reward functional under the stated assumptions; it does not guarantee convergence of clipped GRPO with neural policies.
\end{itemize}

\subsection{Proof of Effective Entropy Gradient}
\label{app:proof_gradient}

\begin{theorem}
Given the Effective Entropy loss $\mathcal{L}_{\text{eff}}(\theta) = \sum_{o \in \mathcal{O}} r(o|q) \pi_\theta(o|q) \log \pi_\theta(o|q)$, its gradient is given by:
\begin{equation}
\nabla_\theta \mathcal{L}_{\text{eff}}(\theta) = \E_{o \sim \pi_\theta} \left[ r(o|q) (1 + \log \pi_\theta(o|q)) \nabla_\theta \log \pi_\theta(o|q) \right]
\end{equation}
\end{theorem}

\begin{proof}
We start by taking the gradient of the loss function with respect to parameters $\theta$. Note that the reward $r(o|q)$ is independent of $\theta$.
\begin{align}
\nabla_\theta \mathcal{L}_{\text{eff}} &= \nabla_\theta \sum_{o \in \mathcal{O}} r(o|q) \pi_\theta(o|q) \log \pi_\theta(o|q) \\
&= \sum_{o \in \mathcal{O}} r(o|q) \nabla_\theta \left( \pi_\theta(o|q) \log \pi_\theta(o|q) \right)
\end{align}
Using the product rule $\nabla(f g) = (\nabla f)g + f(\nabla g)$:
\begin{equation}
\label{eq:product_rule}
\nabla_\theta \left( \pi_\theta \log \pi_\theta \right) = (\nabla_\theta \pi_\theta) \log \pi_\theta + \pi_\theta (\nabla_\theta \log \pi_\theta)
\end{equation}
We apply the log-derivative trick identity: $\nabla_\theta \pi_\theta = \pi_\theta \nabla_\theta \log \pi_\theta$. Substituting this into Eq.~\eqref{eq:product_rule}:
\begin{align}
\nabla_\theta \left( \pi_\theta \log \pi_\theta \right) &= (\pi_\theta \nabla_\theta \log \pi_\theta) \log \pi_\theta + \pi_\theta (\nabla_\theta \log \pi_\theta) \\
&= \pi_\theta (1 + \log \pi_\theta) \nabla_\theta \log \pi_\theta
\end{align}
Substituting this back into the summation:
\begin{align}
\nabla_\theta \mathcal{L}_{\text{eff}} &= \sum_{o \in \mathcal{O}} \pi_\theta(o|q) \left[ r(o|q) (1 + \log \pi_\theta(o|q)) \nabla_\theta \log \pi_\theta(o|q) \right]
\end{align}
By definition of expectation $\E_{x \sim p(x)}[f(x)] = \sum p(x)f(x)$, we obtain the final form:
\begin{equation}
\nabla_\theta \mathcal{L}_{\text{eff}} = \E_{o \sim \pi_\theta} \left[ r(o|q) (1 + \log \pi_\theta(o|q)) \nabla_\theta \log \pi_\theta(o|q) \right]
\end{equation}
\end{proof}

\subsection{Derivation of Token-Level Decomposition}
\label{app:token_decomp}

Here we demonstrate how the sequence-level gradient is decomposed into token-level gradients, justifying the implementation in Eq.~\eqref{eq:token_gradient}.

Let $J(o) = r(o|q)(1 + \log \pi_\theta(o|q))$ be the scalar weight for a sampled trajectory $o$. The gradient estimator for a single sample is:
\begin{equation}
g = J(o) \cdot \nabla_\theta \log \pi_\theta(o|q)
\end{equation}
For an autoregressive model, the sequence log-probability is the sum of token log-probabilities:
\begin{equation}
\log \pi_\theta(o|q) = \sum_{t=1}^T \log \pi_\theta(x_t | x_{<t}, q)
\end{equation}
By the \textbf{linearity of the gradient operator} $\nabla_\theta$, the gradient of a sum is the sum of gradients:
\begin{equation}
\nabla_\theta \log \pi_\theta(o|q) = \nabla_\theta \left( \sum_{t=1}^T \log \pi_\theta(x_t | \dots) \right) = \sum_{t=1}^T \nabla_\theta \log \pi_\theta(x_t | \dots)
\end{equation}
Substituting this back into the estimator $g$:
\begin{align}
g &= J(o) \cdot \left( \sum_{t=1}^T \nabla_\theta \log \pi_\theta(x_t | x_{<t}, q) \right) \\
&= \sum_{t=1}^T \left[ J(o) \cdot \nabla_\theta \log \pi_\theta(x_t | x_{<t}, q) \right]
\end{align}
This proves that applying the sequence-level scalar weight $J(o)$ to the gradient of each token's log-probability is mathematically equivalent to optimizing the sequence-level Effective Entropy. Note that $J(o)$ must be treated as a constant (detached from the computation graph) during the backward pass of the current step.

\section{Posterior Evaluation Metric: Entropy Conversion Efficiency}
\label{app:ece}

While $\mathcal{H}_{\text{eff}}$ serves as an optimization objective, verifying the efficiency of the learning process requires a posterior metric. In standard RL fine-tuning, we typically observe a pattern where the model gains performance ($\Delta \mathcal{P} > 0$) while becoming more confident, leading to a decrease in entropy ($\Delta \mathcal{H} < 0$).

\paragraph{Design Philosophy.}
We conceptualize the reduction of entropy as the ``cost'' or ``fuel'' consumed to acquire performance. Ideally, we seek a metric that represents the \textit{performance gain per unit of entropy cost}, conceptually similar to $\frac{\Delta \mathcal{P}}{|\Delta \mathcal{H}|}$.
However, directly using this ratio introduces significant mathematical issues:
\begin{enumerate}
    \item \textbf{Boundary Instability:} As $\Delta \mathcal{H} \to 0$ (e.g., in the very early stages or during plateaus), the metric approaches infinity, causing numerical explosions that obscure the trend.
    \item \textbf{Sign Ambiguity:} A simple ratio fails to naturally distinguish between beneficial convergence and degenerate collapse without complex conditional logic.
\end{enumerate}
To address this, we propose \textbf{Entropy Conversion Efficiency (ECE)}, which employs an exponential scaling to ensure continuity and stability while preserving the monotonicity of the return-to-cost ratio.
Let $\Delta \mathcal{P}_t = \mathcal{P}_t - \mathcal{P}_{\text{base}}$ and $\Delta \mathcal{H}_t = \mathcal{H}_t - \mathcal{H}_{\text{base}}$. The metric is defined as:
\begin{equation}
\label{eq:ece_def_app}
\text{ECE}_t = \Delta \mathcal{P}_t \cdot \exp\left(\text{sgn}(\Delta \mathcal{P}_t) \cdot \frac{\Delta \mathcal{H}_t}{\mathcal{H}_{\text{base}}} \right)
\end{equation}

\paragraph{Primary Analysis Domain ($\Delta \mathcal{P} > 0, \Delta \mathcal{H} < 0$).}
This is the standard regime for converging models. Here, the term $\frac{\Delta \mathcal{H}}{\mathcal{H}_{\text{base}}}$ is negative.
\begin{itemize}
    \item The ECE value effectively measures the ``exchange rate.'' If two models achieve the same performance gain $\Delta \mathcal{P}$, the one that preserves more entropy (smaller $|\Delta \mathcal{H}|$) will have a larger exponential term (closer to 1), resulting in a higher ECE.
    \item Conversely, a model that collapses (large $|\Delta \mathcal{H}|$) to achieve the same performance is penalized with a smaller exponential factor.
\end{itemize}

\paragraph{Secondary Domains and Caveats.}
While ECE is defined globally, interpretations in other quadrants require caution:
\begin{itemize}
    \item \textbf{Exploratory Growth ($\Delta \mathcal{P} > 0, \Delta \mathcal{H} > 0$):} Mathematically, ECE is amplified here ($\exp > 1$), rewarding simultaneous performance and diversity gains. However, this region is susceptible to ``hacking,'' where a model might artificially inflate entropy (e.g., via irrelevant noise) without genuine reasoning diversity. Therefore, ECE comparisons are most rigorous when $\Delta \mathcal{H} \le 0$.
    \item \textbf{Degeneration ($\Delta \mathcal{P} < 0$):} When performance drops, the $\text{sgn}(\Delta \mathcal{P})$ term inverts the exponent. A collapse in entropy ($\Delta \mathcal{H} < 0$) combined with performance loss results in a heavily penalized (highly negative) score, correctly identifying failure modes.
\end{itemize}
In our experiments, we focus primarily on the efficiency of converting entropy reduction into performance gain, ensuring that the model does not simply memorize solutions (low cost-efficiency) but learns robustly.

\section{Qualitative Analysis of Entropy-aware Baselines}
\label{app:baseline_analysis}

By approximating low probability as High Entropy and high probability as Low Entropy, we can categorize the optimization behaviors into an eight-dimensional table (Table~\ref{tab:entropy_regions}). We perform a qualitative analysis on the following algorithms, filling them into the table based on their contrast with standard GRPO properties. Note that in this context, "Compensation" refers to boosting the update magnitude, while "Penalty" refers to suppressing it.

\begin{table*}[t]
\centering
\caption{Classification of entropy-aware algorithms based on their modification of update magnitudes relative to GRPO. The regions are defined by sample type, entropy level, and update action.}
\label{tab:entropy_regions}
\renewcommand{\arraystretch}{1.25}
\small
\setlength{\tabcolsep}{3pt}
\begin{tabularx}{\textwidth}{cXXXX}
\toprule
\multirow{2}{*}{\textbf{Entropy Level}} 
& \multicolumn{2}{c}{\textbf{Positive Sample ($r=1$)}} 
& \multicolumn{2}{c}{\textbf{Negative Sample ($r=0$)}} \\
\cmidrule(lr){2-3} \cmidrule(lr){4-5}
& \centering\arraybackslash \textbf{Compensate}
& \centering\arraybackslash \textbf{Penalize}
& \centering\arraybackslash \textbf{Compensate}
& \centering\arraybackslash \textbf{Penalize} \\
\midrule
\makecell{\textbf{High Entropy}\\(Low Prob.)}
& \centering\arraybackslash Clip-Higher; Entropy-Reg; Entropy-Adv; GTPO; AEPO; \textbf{VEPO}
& \centering\arraybackslash Adv-Reweighting
& \centering\arraybackslash --
& \centering\arraybackslash Entropy-Reg; Entropy-Adv; Adv-Reweighting; GTPO; Clip-Cov/KL-Cov \\
\midrule
\makecell{\textbf{Low Entropy}\\(High Prob.)}
& \centering\arraybackslash AEPO
& \centering\arraybackslash GTPO; 20/80; Clip-Cov/KL-Cov
& \centering\arraybackslash GTPO
& \centering\arraybackslash 20/80 \\
\bottomrule
\end{tabularx}
\end{table*}

\begin{itemize}[leftmargin=0.5cm]
    \item \textbf{Clip Higher~\citep{yu2025dapo}:} Since clipped tokens are predominantly low-probability tokens~\citep{su2025gppo}, the operation of globally raising the upper bound is equivalent to compensating for the contribution of low-probability (high-entropy) tokens. This belongs to \textbf{Region 1} optimization.
    
    \item \textbf{Entropy Regularization (Entropy-Reg)~\citep{o2016combining}:} According to its gradient formula, it generally improves high-entropy situations but indiscriminately increases probability. Therefore, it compensates for positive samples and penalizes negative samples (by opposing the probability reduction), belonging to \textbf{Region 1 and 4} optimization.
    
    \item \textbf{Entropy Advantage (Entropy-Adv)~\citep{cheng2025reasoning}:} This method performs compensation on positive samples and suppression on negative samples. Since the degree of interference is positively correlated with entropy, it belongs to \textbf{Region 1 and 4} optimization.
    
    \item \textbf{Advantage Reweighting (Adv-Reweighting)~\citep{yang2025not}:} Regardless of whether the sample is positive or negative, the smaller the probability (higher the entropy), the larger the penalty on the gradient. Thus, this is \textbf{Region 2 and 4} optimization.
    
    \item \textbf{GTPO~\citep{tan2025gtpo}:} The optimization direction is similar to Entropy-Adv, but due to the reward weighting mechanism, it also affects low-entropy situations. Low entropy results in penalizing positive sample updates and compensating negative sample updates. Therefore, it spans \textbf{Region 1, 4, 6, and 7} optimization.
    
    \item \textbf{AEPO~\citep{wang2025arbitrary}:} This adds a REINFORCE-like regularization term, which can be viewed entirely as treating samples as positive. Thus, it falls into \textbf{Region 1 and 5} optimization.
    
    \item \textbf{20/80 Strategy~\citep{wang2025beyond}:} This method stops gradients for low-entropy tokens. This equates to suppressing the update for both positive and negative samples when entropy is low, belonging to \textbf{Region 6 and 8} optimization.
    
    \item \textbf{Clip-Cov \& KL-Cov~\citep{cui2025entropy} :} These methods suppress high-covariance tokens. Based on the analysis, these naturally fall into \textbf{Region 4 and 6} optimization.
    
    \item \textbf{VEPO (Ours):} Our method specifically compensates for high entropy in positive samples, strictly belonging to \textbf{Region 1} optimization.
\end{itemize}

\section{Implementation Details}
\label{app:implementation_details}

\subsection{Infrastructure and Environment}
\paragraph{Computational Resources.}
Our experiments were conducted on a Meituan-provided high-performance computing cluster comprising 30 nodes. Each node is equipped with 8 $\times$ NVIDIA A100 GPUs (80GB VRAM), interconnected via NVLink for high-bandwidth intra-node communication and InfiniBand for inter-node synchronization. The total compute budget for the experimental sweep, including hyperparameter tuning and baseline reproduction, exceeded 25,000 GPU-hours.

\paragraph{Software Stack.}
We implemented VEPO within the \textbf{VeRL} framework~\citep{sheng2025hybridflow}, a highly efficient library designed for Post-Training Large Language Models.
\begin{itemize}
    \item \textbf{Training Backend:} We utilized Fully Sharded Data Parallel (FSDP) to manage memory efficiency. For the 7B models, we employed ZeRO-3 offloading to enable full-parameter fine-tuning without requiring parameter-efficient approximations (PEFT).
    \item \textbf{Inference Engine:} To maximize the throughput of the group-based rollout generation, we integrated \textbf{vLLM} as the inference backend. We leveraged the PagedAttention mechanism to efficiently manage Key-Value (KV) cache memory, which is critical when generating long chain-of-thought sequences (up to 8192 tokens) to prevent OOM errors during the exploration phase.
\end{itemize}

\subsection{Data and Evaluation Benchmarks}
\paragraph{Training Dataset.}
We utilized the \textbf{DAPO-Math-17k} dataset~\citep{yu2025dapo} for training. To ensure strict alignment between the pre-trained backbone and the RL fine-tuning stage, we used the official tokenizer from the Qwen2.5-Math series. We enforced a maximum sequence length of 8192 tokens, allowing the model sufficient capacity to explore complex, multi-step reasoning paths without artificial truncation.

\paragraph{Reward Function.}
We implemented a strict dual-oracle verification protocol to prevent reward hacking and ensure metric validity:
\begin{itemize}
    \item \textbf{Training Oracle:} We utilized the lightweight \texttt{math\_verify} library for efficient online reward calculation. The reward is binary: $r=1$ if the boxed answer matches the ground truth, and $r=0$ otherwise.
    \item \textbf{Evaluation Oracle:} For all reported metrics, we employed the more rigorous \texttt{prime\_math} evaluation suite~\citep{lightman2023let}. This ensures that our reported gains are not artifacts of loose parsing logic.
\end{itemize}

\paragraph{Benchmark Suite.}
\label{app:benchmarks}
Table \ref{tab:benchmark_details} details the characteristics of the evaluation datasets.

\begin{table*}[t]
\centering
\small
\renewcommand{\arraystretch}{1.15}
\setlength{\tabcolsep}{4pt}
\caption{Detailed profile of the reasoning benchmarks. Datasets are grouped by their strategic role in evaluating the model's latent reasoning capabilities.}
\label{tab:benchmark_details}
\begin{tabularx}{\textwidth}{l>{\raggedright\arraybackslash}p{2.8cm}X}
\toprule
\textbf{Category} & \textbf{Dataset} & \textbf{Strategic Evaluation Focus} \\
\midrule
\rowcolor{gray!10}
\multicolumn{3}{l}{\textit{Competition-Level Reasoning (High-Stakes Deduction)}} \\
& \textbf{AIME 2024/2025} 
& Targets the robustness of long-range reasoning chains; the 2025 version serves as a temporal hold-out to reduce the risk of data leakage. \\
& \textbf{AMC 2023} 
& Benchmarks the model against standard competitive mathematics at the entry olympiad level. \\
\midrule
\rowcolor{gray!10}
\multicolumn{3}{l}{\textit{Broad-Spectrum Knowledge (Domain Versatility)}} \\
& \textbf{MATH-500} 
& Provides a curated cross-section of scholastic math to evaluate whether the policy maintains balanced performance across major mathematical sub-fields. \\
& \textbf{Minerva} 
& Focuses on the transition from standard mathematical problems to complex, scientific-grade symbolic reasoning. \\
\midrule
\rowcolor{gray!10}
\multicolumn{3}{l}{\textit{Stress Testing (Performance Ceiling)}} \\
& \textbf{OlympiadBench} 
& Challenges the model with rigorous olympiad-level problems to probe the limits of validity-aware exploration. \\
\bottomrule
\end{tabularx}
\end{table*}

\subsection{Training Protocols}
\paragraph{Optimization Strategy.}
We employed the AdamW optimizer with $\beta_1=0.9$, $\beta_2=0.95$, and a weight decay of $0.1$. To strictly isolate the algorithmic contribution of VEPO, we adopted a \textbf{constant learning rate} strategy across all experiments, avoiding the confounding effects of warm-up or cosine decay schedules.
\begin{itemize}
    \item \textbf{Batch Configuration:} The global batch size is 512 and the mini-batch size is 32.
    \item \textbf{KL Divergence:} We set the KL penalty coefficient $\beta_\text{KL}=0$. We rely implicitly on the trust region constraints imposed by the PPO clipping mechanism ($\epsilon=0.2$) to prevent policy collapse.
\end{itemize}

\subsection{Hyperparameter Specifications}
\label{app:hyperparameters}

We performed a grid search for the Effective Entropy coefficient $\beta$ on a held-out validation set. Table \ref{tab:hyper_settings} lists the optimal configurations used for the main results, along with the specific settings for all baseline algorithms.

\begin{table*}[h!]
    \centering
    \small
    \caption{Detailed hyperparameter configurations for VEPO and all baselines. For baselines, we utilized the values recommended in their respective original papers or tuned them for optimal performance on our validation set.}
    \renewcommand{\arraystretch}{1.25}
    \begin{tabularx}{\textwidth}{l l c >{\raggedright\arraybackslash}X}
    \toprule
    \textbf{Model Scale} & \textbf{Algorithm} & \textbf{Learning Rate} & \textbf{Key Hyperparameters} \\
    \midrule
    \multicolumn{4}{p{\dimexpr\textwidth-2\tabcolsep\relax}}{\textit{Common Settings: Global Batch Size = 512, Mini-batch Size = 32, Max Length = 8192, $\beta_\text{KL}=0$}} \\
    \midrule
    \multirow{7}{*}{\textbf{1.5B}} 
     & GRPO & \multirow{7}{*}{$1.0 \times 10^{-6}$} & $\varepsilon=0.2$ \\
     & Entropy-Reg & & $\beta_\text{reg}=0.001$ \\
     & Entropy-Adv & & $\alpha_\text{E}=0.4, \kappa=2.0$ \\
     & Adv-Reweighting & & $\alpha_\text{A}=0.1$ \\
     & GTPO & & $\alpha_1=1.0, \alpha_2=0.1$ \\
     & AEPO & & $\mathcal{H}_\text{target}=0.75, T_{\text{low}}=0.8, T_{\text{high}}=1.2, \alpha=0.1$ \\
     & \textbf{VEPO} & & $\beta=0.0001$ \\
    \midrule
    \multirow{7}{*}{\textbf{7B}} 
     & GRPO & \multirow{7}{*}{$4.63 \times 10^{-7}$} & $\varepsilon=0.2$ \\
     & Entropy-Reg & & $\beta_\text{reg}=0.001$ \\
     & Entropy-Adv & & $\alpha_\text{E}=0.4, \kappa=2.0$ \\
     & Adv-Reweighting & & $\alpha_\text{A}=0.1$ \\
     & GTPO & & $\alpha_1=1.0, \alpha_2=0.1$ \\
     & AEPO & & $\mathcal{H}_\text{target}=0.75, T_{\text{low}}=0.8, T_{\text{high}}=1.2, \alpha=0.1$ \\
     & \textbf{VEPO} & & $\beta=0.0001$ \\
    \bottomrule
    \end{tabularx}
    \label{tab:hyper_settings}
\end{table*}

\section{Additional Experimental Results}
\label{app:additional_results}

\subsection{The Paradox of Unaligned Diversity}
\label{app:entropy_paradox}

\begin{figure*}[t]
\centering
\begin{subfigure}{0.24\columnwidth}
\centering
\includegraphics[width=\linewidth]{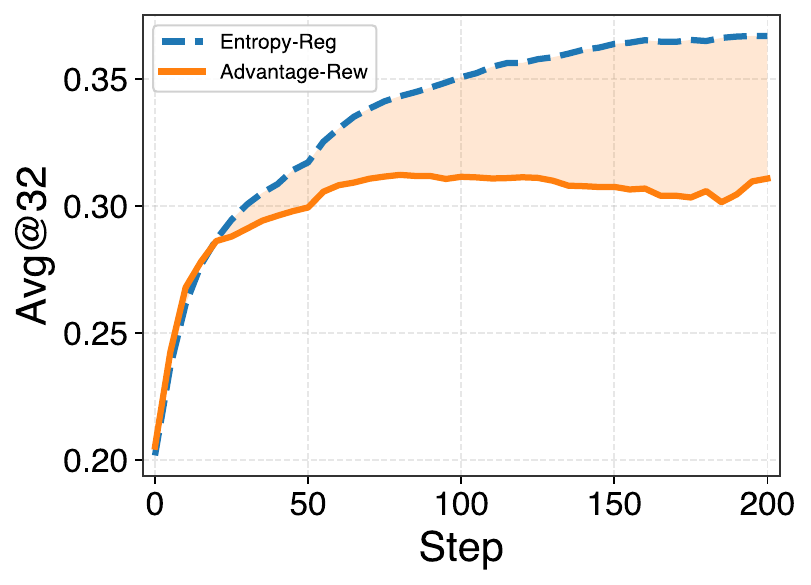}
\caption{AIME25 Avg@32}
\end{subfigure}
\hfill
\begin{subfigure}{0.24\columnwidth}
\centering
\includegraphics[width=\linewidth]{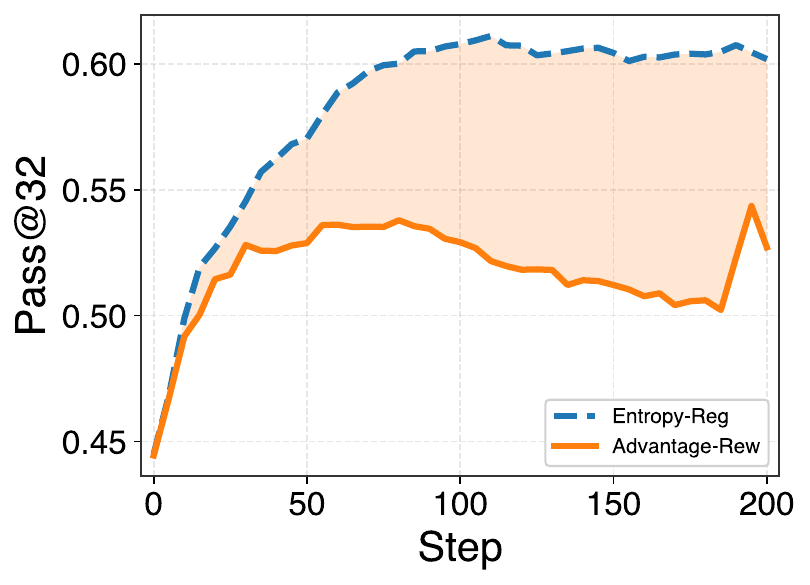}
\caption{AIME25 Pass@32}
\end{subfigure}
\hfill
\begin{subfigure}{0.24\columnwidth}
\centering
\includegraphics[width=\linewidth]{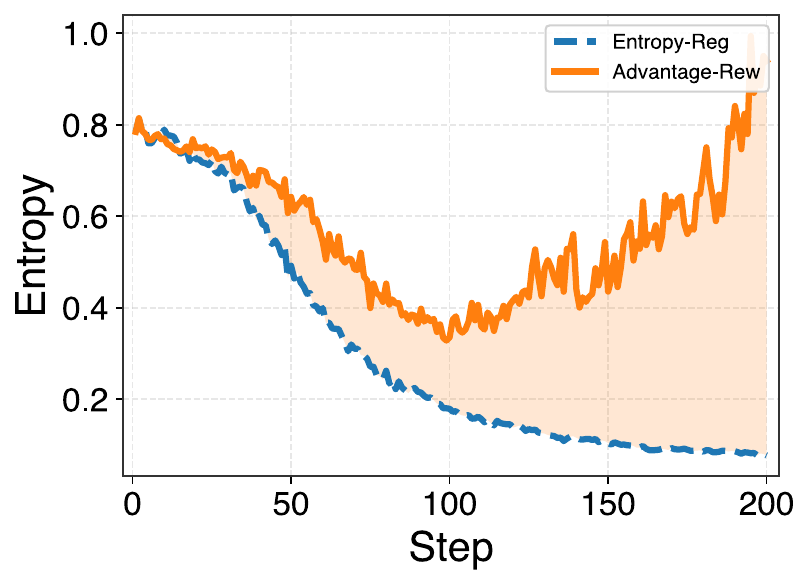}
\caption{Entropy ($\mathcal{H}_{\text{tok}}$)}
\end{subfigure}
\hfill
\begin{subfigure}{0.24\columnwidth}
\centering
\includegraphics[width=\linewidth]{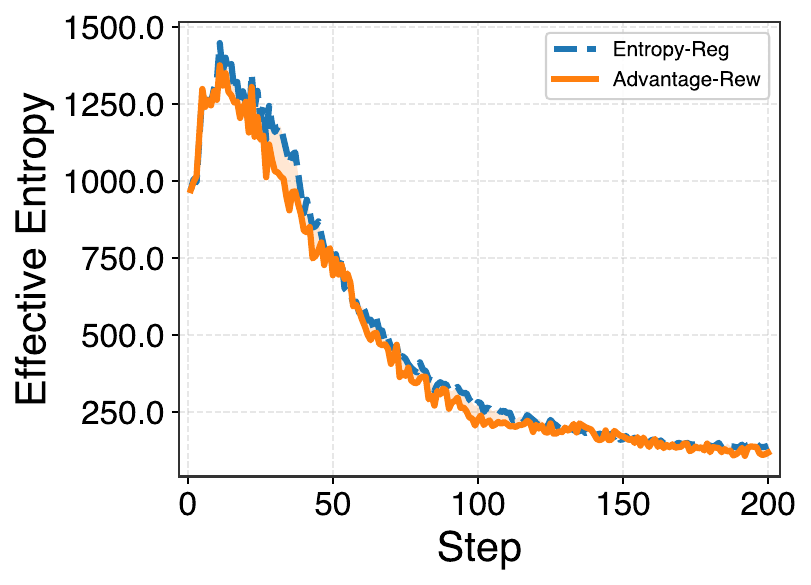}
\caption{Effective Entropy}
\end{subfigure}
\caption{Comparative dynamics between Entropy-Reg and Adv-Reweighting on 7B model. While Adv-Reweighting maintains or even surges in total entropy, its effective entropy remains consistently below Entropy-Reg, resulting in inferior reasoning performance.}
\label{fig:app_paradox}
\end{figure*}

To further illustrate the necessity of aligning exploration with validity, we provide a comparative analysis of the training dynamics between \textbf{Entropy-Reg} and \textbf{Adv-Reweighting}. As shown in Figure~\ref{fig:app_paradox}, these two baselines exhibit a phenomenon we term the \textit{"Diversity Paradox"}.

\paragraph{High Entropy Without Validity.} 
In the late stages of training (beyond step 150), Adv-Reweighting exhibits a sharp surge in token-level entropy ($ \mathcal{H}_{\text{tok}} $). Under traditional metrics, this would be interpreted as a more "diverse" or "exploratory" model. However, this diversity fails to translate into performance; in fact, its Avg@32 and Pass@32 metrics plateau or even decline during this period. 

\paragraph{Mechanism Failure.} 
The root cause lies in the \textit{Effective Entropy}. Despite having higher total uncertainty, Adv-Reweighting's effective entropy is lower than that of the simpler Entropy-Reg. This indicates that the additional entropy in Adv-Reweighting is being injected into the \textbf{invalid subspace}. Without VEPO's validity-aware attraction mechanism, the model generates diverse hallucinations (high total entropy) rather than diverse correct solutions. This confirms that unaligned entropy regularization can be detrimental, acting as a source of chaotic noise that distracts the model from the manifold of correct reasoning.

\subsection{Training Efficiency and Computational Overhead}
\label{app:overhead}

Matched training diagnostics show a 9.8\% steady-state overhead for VEPO relative to GRPO. VEPO reaches the paired GRPO run's final six-benchmark aggregate Avg@32 after 73\% of its rollout budget at 1.5B and 78\% at 7B. Including the measured overhead gives approximately 80\% and 86\% of GRPO's wall-clock time to that target. These are time-to-target measurements for the tested runs, not evidence of zero computational cost.

With $G=8$, the fractions of rollouts carrying a nonzero auxiliary signal in the early, middle, and late thirds of training are 28.6\%, 39.7\%, and 45.1\% at 1.5B, and 42.1\%, 51.5\%, and 56.6\% at 7B. The corresponding fractions of groups with no auxiliary signal are 7.1\%, 2.0\%, and 0.9\%, versus 1.3\%, 0.4\%, and 0.1\%. The median auxiliary-to-GRPO gradient-norm ratios, including the tuned $\beta$, are 0.18 and 0.15. Incorrect trajectories in mixed-reward groups retain the ordinary GRPO update; they receive no additional effective-entropy signal. Terminal verification nevertheless cannot assign credit to useful intermediate steps in an outcome-wrong trajectory.

\section{Selected Additional Evidence}
\label{app:selected_evidence}

\subsection{Strategy-Diversity Robustness}
\label{app:diversity_checks}

\paragraph{Matching Correct Samples per Problem.}
Pass@$k$ measures repeated-sampling success, which also depends on per-sample accuracy. We therefore measure strategy diversity directly on MATH500 with the 7B backbone. We retain the 441 of 500 problems on which both GRPO and VEPO produce at least eight answer-verified correct trajectories in their fixed pools of 32 samples. On each problem, we draw exactly eight correct trajectories per method without replacement and average over 10,000 rarefactions with fixed random seeds. This controls the number of correct samples separately for every problem, rather than only matching dataset-level accuracy.

We strip final-answer spans, pool the remaining chains of thought from both methods, and cluster within each problem using frozen gte-Qwen2-7B-instruct embeddings~\citep{gteqwen2024}, L2 normalization, and average-linkage agglomerative clustering with cosine distance~\citep{mullner2011modern}. A strategy denotes a shared mathematical plan, invariant, decomposition, or substitution, irrespective of wording or notation. Before comparing methods, we fix the pipeline and select one global distance threshold on method-blinded development annotations. Two mathematical annotators label 600 same/different-strategy pairs with Cohen's $\kappa=0.82$; the frozen pipeline achieves precision 0.88 and recall 0.84 on a disjoint held-out set.

Two internal university faculty members provided unpaid, method-blinded annotations of reasoning-strategy equivalence.

The protocol in Section~\ref{sec:direct_diversity} fixes its threshold by development-set pairwise F1 before comparing methods. For cluster proportions $s_j$, Hill-1 modes are $\exp(-\sum_j s_j\log s_j)$; the largest-cluster share is $\max_j s_j$. Table~\ref{tab:diversity_robustness} retains the matched count of eight trajectories per method per problem. Length matching is one-to-one within a problem without replacement.

\begin{center}
\begin{minipage}{\linewidth}
\centering
\small
\captionof{table}{Robustness of the paired VEPO-minus-GRPO cluster-count difference on MATH500-7B. The matched correct count is eight trajectories per method per problem.}
\label{tab:diversity_robustness}
\begin{tabular}{lcc}
\toprule
Protocol & Difference & 95\% confidence interval \\
\midrule
Primary gte-Qwen2 encoder & $+0.37$ & $[0.24,0.50]$ \\
Prespecified e5-Mistral encoder~\citep{wang2024embeddings} & $+0.31$ & $[0.18,0.44]$ \\
Within-problem CoT-length matching & $+0.34$ & $[0.20,0.47]$ \\
\bottomrule
\end{tabular}
\end{minipage}
\end{center}

A separate blinded rationale audit finds 92.5\% of GRPO and 93.0\% of VEPO correct-answer trajectories logically sound: difference $+0.5$ percentage points, 95\% CI $[-3.7,4.7]$. Answer verification therefore does not certify every intermediate step. The diversity endpoint also conditions on the 441 commonly eligible problems, rather than all MATH500 problems.

\subsection{Training-Seed Variability}
\label{app:statistical_details}

GRPO and VEPO were trained with the same three paired random seeds, $0$, $1$, and $2$.

\begin{center}
\begin{minipage}{\linewidth}
\centering
\small
\captionof{table}{VEPO minus GRPO across three paired training seeds, in percentage points, averaged over the six mathematical benchmarks.}
\label{tab:paired_seeds}
\begin{tabular}{lcc}
\toprule
Backbone & Avg@32 gains & Pass@32 gains \\
\midrule
1.5B & $+2.1,\ +2.4,\ +2.1$ & $+1.3,\ +1.6,\ +1.6$ \\
7B & $+2.3,\ +2.4,\ +1.9$ & $+1.3,\ +1.4,\ +1.2$ \\
\bottomrule
\end{tabular}
\end{minipage}
\end{center}

Aggregate Avg@32 is $48.4\pm0.46$ versus $50.6\pm0.60$ at 1.5B and $58.9\pm0.44$ versus $61.1\pm0.40$ at 7B (GRPO versus VEPO; $\pm$ denotes seed SD). These three training replicates are distinct from the problem-bootstrap uncertainty in the diversity evaluation. Against AEPO, the 7B aggregate Pass@32 difference is $+0.4$ percentage points with reported 95\% interval $[-0.3,1.1]$, supporting competitiveness rather than significant superiority.

\subsection{Sequence-Weight Controls}
\label{app:weight_controls}

The constant-gated control preserves the correctness mask, attraction direction, raw sequence sum, and held-out tuning budget while removing the $\log\pi$ dependence. Its three paired Avg@32 deficits relative to VEPO are $1.2,1.5,1.4$ percentage points at 1.5B and $1.1,1.3,1.2$ at 7B. Response-length normalization instead divides each trajectory's contribution to $\mathcal H_{\mathrm{eff}}$ by $T(o)$; it defines a distinct functional and supplies no causal token credit.

\begin{center}
\begin{minipage}{\linewidth}
\centering
\small
\captionof{table}{Sequence-weight controls. Values are unweighted Avg@32 / Pass@32 aggregates over the six mathematical benchmarks in the original-run setting. Length normalization is an additional variant, not the objective used in the main experiments.}
\label{tab:weight_controls}
\begin{tabular}{lcc}
\toprule
Variant & 1.5B & 7B \\
\midrule
Constant-gated sequence sum & 49.2 / 70.8 & 59.9 / 77.7 \\
VEPO, raw effective entropy & 50.6 / 71.6 & 61.1 / 78.5 \\
Response-length-normalized VEPO & -- & 61.2 / 78.6 \\
\bottomrule
\end{tabular}
\end{minipage}
\end{center}

\subsection{Controlled Verifier Noise}
\label{app:verifier_noise}

For constant class-conditional false-negative rate $\alpha$ and false-positive rate $\gamma$,
\begin{equation}
\mathbb E[\widetilde{\mathcal H}_{\mathrm{eff}}]
=\gamma\mathcal H_{\mathrm{seq}}+(1-\alpha-\gamma)\mathcal H_{\mathrm{eff}}.
\end{equation}
False negatives suppress valid-space exploration, while false positives reintroduce an entropy incentive on incorrect outputs. This simplification does not apply to arbitrary trajectory-dependent verifier errors. Table~\ref{tab:verifier_noise} shows retained VEPO gains within the tested noise range and degradation of both methods as noise increases.

\begin{center}
\begin{minipage}{\linewidth}
\centering
\small
\captionof{table}{Controlled verifier-noise retraining on the 1.5B backbone. Values are three-seed means of six-benchmark Avg@32 / Pass@32. FN and FP denote injected false-negative and false-positive noise.}
\label{tab:verifier_noise}
\begin{tabular}{lcc}
\toprule
Verifier setting & GRPO & VEPO \\
\midrule
Clean & 48.4 / 70.1 & 50.6 / 71.6 \\
FN 5\% & 48.2 / 69.8 & 50.3 / 71.2 \\
FN 10\% & 47.8 / 69.3 & 49.9 / 70.8 \\
FP 5\% & 47.6 / 68.9 & 49.6 / 70.4 \\
FP 10\% & 46.8 / 67.7 & 48.8 / 69.6 \\
\bottomrule
\end{tabular}
\end{minipage}
\end{center}

\paragraph{Non-binary Rewards.}
Replacing the binary gate by a score $v(o)\in[0,1]$ gives reward-weighted entropy, $-\sum_o v(o)\pi(o)\log\pi(o)$, rather than entropy restricted to a binary valid subspace. Partial correctness and process-level verification remain extensions.

\subsection{Valid Mass and Conditional Diversity}
\label{app:valid_mass}
Effective entropy is an unnormalized partial entropy, not the entropy of the conditional distribution over valid outputs. Writing $Z=\sum_{r(o)=1}\pi_\theta(o\mid q)$ and $p_{\mathrm{valid}}(o)=\pi_\theta(o\mid q)/Z$ for $Z>0$ gives
\begin{equation}
\mathcal H_{\mathrm{eff}}
=Z\,\mathcal H(p_{\mathrm{valid}})-Z\log Z.
\end{equation}
It therefore depends on both valid probability mass and diversity within that mass. Section~\ref{sec:direct_diversity} evaluates strategy diversity separately with matched correct counts. Here validity means terminal-verifier acceptance, which need not certify every intermediate reasoning step.

\subsection{Detached-Weight Loss Pseudocode}
\label{app:loss_pseudocode}

The following pseudocode implements the auxiliary gradient in Eq.~\ref{eq:token_gradient} on a batch of sampled responses. The mask excludes prompts and padding, and includes the response tokens used in the sequence probability. The raw sequence sum is retained; taking a token mean here would change the stated objective. The loss has the minimization sign, and detaching the weight avoids differentiating it twice.

\begin{quote}
\begin{verbatim}
# logp: current-policy response token log probabilities [B, T]
# mask: response-token mask; reward: binary terminal reward [B]
sequence_logp = (logp * mask).sum(dim=-1)
weight = (reward * (1.0 + sequence_logp)).detach()
auxiliary_loss = (weight * sequence_logp).mean()
loss = grpo_loss + beta * auxiliary_loss
loss.backward()
\end{verbatim}
\end{quote}

This is a computational surrogate whose gradient matches the derived on-policy estimator at the sampled policy. Reusing a batch from an old policy does not by itself make it an unbiased estimator under the updated policy. Exact current-policy expectation would require fresh samples or appropriate sequence-level importance weighting; the original VEPO update is kept as the simple local surrogate.

\section{Limitations}
\label{app:limitations}

This work has several limitations. First, effective entropy relies on a reliable validity signal, and is therefore most directly applicable to RLVR settings where correctness can be verified by exact answers, unit tests, program execution, or other deterministic reward functions. Extending the same idea to open-ended tasks with ambiguous or preference-based rewards requires a more careful definition of the valid subspace. Second, VEPO is intentionally designed as a simple instantiation of effective entropy rather than a highly engineered optimizer. Although this simplicity helps isolate the contribution of the concept, more adaptive variants may further improve stability and efficiency. Third, our experiments focus on mathematical reasoning and models up to the tested scales. Whether the same gains hold for larger models, multimodal reasoning, tool-use agents, or long-horizon interactive tasks remains to be further validated. Finally, effective entropy is estimated from sampled rollouts, so its quality depends on sampling coverage: when valid trajectories are extremely rare in early training, additional warm-up, curriculum design, or stronger initial policies may be needed.

The direct strategy-diversity result conditions on commonly solved prompts and on a finite sample budget; embedding clusters are an operational proxy for strategies, not a proof of complete solution-space coverage. Answer acceptance also does not ensure every intermediate step is logically sound. False-positive verifier errors can encourage incorrect outputs, and sequence-level weights do not solve causal token-level credit assignment. The three-seed evidence supports consistency in the tested configuration, with limited precision for seed-level inference.

\section{Broader Impacts}
\label{app:broader_impacts}

Validity-aware exploration may improve the reliability of mathematical reasoning systems by favoring answer-verified outputs. A terminal verifier can still accept a correct answer accompanied by an unsound rationale, so downstream educational or scientific uses require independent checks. Poorly specified or malicious verifiers could reward harmful behavior; improving task correctness does not establish general safety. The method also requires repeated rollouts and therefore incurs substantial compute cost. We use existing mathematical datasets and do not collect private user data.

\clearpage
\section*{NeurIPS Paper Checklist}

\begin{enumerate}

\item {\bf Claims}
    \item[] Question: Do the main claims made in the abstract and introduction accurately reflect the paper's contributions and scope?
    \item[] Answer: \answerYes{}
    \item[] Justification: The claims are scoped to mathematical RLVR on the tested 1.5B and 7B backbones. Aggregate accuracy and direct matched-count strategy diversity are reported in Sections~\ref{sec:main_results} and~\ref{sec:direct_diversity}; the complete paired-seed results appear in Appendix~\ref{app:statistical_details}.

\item {\bf Limitations}
    \item[] Question: Does the paper discuss the limitations of the work performed by the authors?
    \item[] Answer: \answerYes{}
    \item[] Justification: The paper discusses limitations in Appendix~\ref{app:limitations}. It explicitly notes the dependence on reliable validity signals, the intentionally simple design of VEPO, the current experimental focus on mathematical reasoning and tested model scales, and the sampling-coverage issue when valid trajectories are rare in early training.

\item {\bf Theory assumptions and proofs}
    \item[] Question: For each theoretical result, does the paper provide the full set of assumptions and a complete (and correct) proof?
    \item[] Answer: \answerYes{}
    \item[] Justification: The paper states the binary-verifier setting and derives the on-policy entropy gradient. Appendix~\ref{app:optima_analysis} restricts the finite-simplex optimum to the stated coefficient condition; the local reused-rollout surrogate is distinguished from an exact current-policy estimator.

    \item {\bf Experimental result reproducibility}
    \item[] Question: Does the paper fully disclose all the information needed to reproduce the main experimental results of the paper to the extent that it affects the main claims and/or conclusions of the paper (regardless of whether the code and data are provided or not)?
    \item[] Answer: \answerNo{}
    \item[] Justification: The paper provides the objective, pseudocode, models, training data, key hyperparameters, and paired training seeds ($0,1,2$). The implementation and executable reproduction scripts are not publicly released.

\item {\bf Open access to data and code}
    \item[] Question: Does the paper provide open access to the data and code, with sufficient instructions to faithfully reproduce the main experimental results, as described in supplemental material?
    \item[] Answer: \answerNo{}
    \item[] Justification: The existing datasets, models, and encoder are cited. The authors are not publicly releasing the VEPO implementation or reproduction scripts with this version.

\item {\bf Experimental setting/details}
    \item[] Question: Does the paper specify all the training and test details (e.g., data splits, hyperparameters, how they were chosen, type of optimizer) necessary to understand the results?
    \item[] Answer: \answerNo{}
    \item[] Justification: Section~\ref{sec:experimental_setup} and Appendix~\ref{app:implementation_details} describe the models, datasets, optimizer, group size, clipping, and coefficient selection. Appendix~\ref{app:statistical_details} reports paired training seeds $0,1,2$; Appendix~\ref{app:diversity_checks} describes the strategy-diversity evaluation and threshold-selection procedure.

\item {\bf Experiment statistical significance}
    \item[] Question: Does the paper report error bars suitably and correctly defined or other appropriate information about the statistical significance of the experiments?
    \item[] Answer: \answerYes{}
    \item[] Justification: Section~\ref{sec:direct_diversity} reports a problem-bootstrap confidence interval and paired sign-flip permutation test for the primary diversity endpoint. Table~\ref{tab:paired_seeds} reports all three paired seed deltas; Appendix~\ref{app:statistical_details} distinguishes seed standard deviations from evaluation uncertainty.

\item {\bf Experiments compute resources}
    \item[] Question: For each experiment, does the paper provide sufficient information on the computer resources (type of compute workers, memory, time of execution) needed to reproduce the experiments?
    \item[] Answer: \answerYes{}
    \item[] Justification: Appendix~\ref{app:implementation_details} describes the Meituan-provided A100 hardware and total experimental compute budget. Appendix~\ref{app:overhead} reports matched steady-state overhead and rollout and wall-clock time to a fixed accuracy target.
    
\item {\bf Code of ethics}
    \item[] Question: Does the research conducted in the paper conform, in every respect, with the NeurIPS Code of Ethics \url{https://neurips.cc/public/EthicsGuidelines}?
    \item[] Answer: \answerYes{}
    \item[] Justification: The research uses existing mathematical datasets, cited model backbones, automatic verifiers, and blinded mathematical annotation. It does not collect private user data. Expert annotation was performed by internal university faculty without monetary compensation.

\item {\bf Broader impacts}
    \item[] Question: Does the paper discuss both potential positive societal impacts and negative societal impacts of the work performed?
    \item[] Answer: \answerYes{}
    \item[] Justification: Appendix~\ref{app:broader_impacts} discusses reliability benefits, unsound rationales accepted by terminal verifiers, verifier misuse, and computational cost.
    
\item {\bf Safeguards}
    \item[] Question: Does the paper describe safeguards that have been put in place for responsible release of data or models that have a high risk for misuse (e.g., pre-trained language models, image generators, or scraped datasets)?
    \item[] Answer: \answerNA{}
    \item[] Justification: The paper does not release a new pretrained language model, scraped dataset, image generator, or other high-risk asset. The work proposes an optimization method and evaluates it on mathematical reasoning benchmarks using existing models and datasets.

\item {\bf Licenses for existing assets}
    \item[] Question: Are the creators or original owners of assets (e.g., code, data, models), used in the paper, properly credited and are the license and terms of use explicitly mentioned and properly respected?
    \item[] Answer: \answerNo{}
    \item[] Justification: The paper properly cites the existing models, datasets, frameworks, and verification tools used in the experiments. However, the current version does not explicitly list the license or terms of use for each asset.

\item {\bf New assets}
    \item[] Question: Are new assets introduced in the paper well documented and is the documentation provided alongside the assets?
    \item[] Answer: \answerNA{}
    \item[] Justification: The paper does not introduce or release a new dataset, benchmark, or pretrained model as a new asset. The main contribution is a training objective and optimization method built on existing models and datasets.

\item {\bf Crowdsourcing and research with human subjects}
    \item[] Question: For crowdsourcing experiments and research with human subjects, does the paper include the full text of instructions given to participants and screenshots, if applicable, as well as details about compensation (if any)? 
    \item[] Answer: \answerNo{}
    \item[] Justification: Appendix~\ref{app:diversity_checks} summarizes the method-blinded strategy-equivalence annotation and reports validation results. Two internal university faculty members performed the annotation without monetary compensation.

\item {\bf Institutional review board (IRB) approvals or equivalent for research with human subjects}
    \item[] Question: Does the paper describe potential risks incurred by study participants, whether such risks were disclosed to the subjects, and whether Institutional Review Board (IRB) approvals (or an equivalent approval/review based on the requirements of your country or institution) were obtained?
    \item[] Answer: \answerNo{}
    \item[] Justification: The diversity evaluation uses internal university faculty as expert annotators. This version does not document an institutional determination that this expert annotation is exempt from or outside the scope of human-subjects review, or any applicable approval. No exemption or approval is claimed.

\item {\bf Declaration of LLM usage}
    \item[] Question: Does the paper describe the usage of LLMs if it is an important, original, or non-standard component of the core methods in this research? Note that if the LLM is used only for writing, editing, or formatting purposes and does \emph{not} impact the core methodology, scientific rigor, or originality of the research, declaration is not required.
    \item[] Answer: \answerYes{}
    \item[] Justification: LLMs are central to the research, as the paper studies RLVR post-training of DeepSeek-R1-Distill-Qwen backbones and evaluates their mathematical reasoning behavior. The use of LLMs as the experimental models is described in Section~\ref{sec:experimental_setup} and Appendix~\ref{app:implementation_details}.

\end{enumerate}

\end{document}